\documentclass[sigconf]{acmart}

\AtBeginDocument{%
  \providecommand\BibTeX{{%
    \normalfont B\kern-0.5em{\scshape i\kern-0.25em b}\kern-0.8em\TeX}}}

\copyrightyear{2024}
\acmYear{2024}
\setcopyright{acmlicensed}
\acmConference[KDD '24]{Proceedings of the 30th ACM SIGKDD Conference on Knowledge Discovery and Data Mining}{August 25--29, 2024}{Barcelona, Spain}
\acmBooktitle{Proceedings of the 30th ACM SIGKDD Conference on Knowledge Discovery and Data Mining (KDD '24), August 25--29, 2024, Barcelona, Spain} \acmDOI{10.1145/3637528.3671507}
\acmISBN{979-8-4007-0490-1/24/08}

\usepackage{algorithm}
\usepackage{algorithmic}
\usepackage{xspace}
\usepackage{multirow}
\usepackage{multicol}
\usepackage{graphicx}  
\usepackage{float}  
\usepackage{subfigure}  
\usepackage{pifont}

\newtheorem{myDef}{Definition}
\newtheorem{prob}{Problem}
\newtheorem{assumption}{Assumption}

\newcommand{\etal}{\emph{et~al.}}
\newcommand{\eg}{\emph{e.g.},\/~}

\newcommand{\ie}{\emph{i.e.},\/~}
\newcommand{\wrt}{\emph{w.r.t.}\/~}

\newcommand\figref[1]{Figure~\ref{#1}}

\newcommand\tabref[1]{Table~\ref{#1}}
\newcommand\secref[1]{Section~\ref{#1}}
\newcommand\appref[1]{Appendix~\ref{#1}}
\newcommand\equref[1]{Equation~(\ref{#1})}
\newcommand{\model}{{\sc CP-MoE}\xspace}

\newcommand{\eat}[1]{}

\newcommand{\TODO}[1]{{\color{red}TODO: {#1}}}

\newcommand{\rev}[1]{{\color{purple}#1}} 
\newcommand{\wz}[1]{{\color[rgb]{0.5,0.7,0.4}{#1}}}

\begin{document}




\title{Interpretable Cascading Mixture-of-Experts for Urban Traffic Congestion Prediction}


\author{Wenzhao Jiang}
\orcid{0009-0006-1081-8684}
\affiliation{
\institution{The Hong Kong University of Science and Technology (Guangzhou)}
\city{Guangzhou}
\state{Guangdong}
\country{China}
}
\email{wjiang431@connect.hkust-gz.edu.cn}

\author{Jindong Han}
\orcid{0000-0002-1542-6149}
\affiliation{
\institution{The Hong Kong University of Science and Technology}
\city{Hong Kong}
\country{China}
}
\email{jhanao@connect.ust.hk}

\author{Hao Liu}
\authornote{Corresponding author.}
\orcid{0000-0003-4271-1567}
\affiliation{
\institution{The Hong Kong University of Science and Technology (Guangzhou)}
\city{Guangzhou}
\state{Guangdong}
\country{China}
}
\affiliation{
\institution{The Hong Kong University of Science and Technology}
\city{Hong Kong}
\country{China}
}
\email{liuh@ust.hk}

\author{Tao Tao}
\orcid{0009-0002-1378-3422}
\affiliation{
\institution{Didichuxing Co. Ltd}
\city{Beijing}
\country{China}
}
\email{taotao@didiglobal.com}

\author{Naiqiang Tan}
\orcid{0009-0008-4687-5212}
\affiliation{
\institution{Didichuxing Co. Ltd}
\city{Beijing}
\country{China}
}
\email{tannaiqiang@didiglobal.com}

\author{Hui Xiong}
\orcid{0000-0001-6016-6465}
\affiliation{
\institution{The Hong Kong University of Science and Technology (Guangzhou)}
\city{Guangzhou}
\state{Guangdong}
\country{China}
}
\affiliation{
\institution{The Hong Kong University of Science and Technology}
\city{Hong Kong}
\country{China}
}
\email{xionghui@ust.hk}


\begin{abstract}

Rapid urbanization has significantly escalated traffic congestion, underscoring the need for advanced congestion prediction services to bolster intelligent transportation systems.
As one of the world's largest ride-hailing platforms, DiDi places great emphasis on the accuracy of congestion prediction to enhance the effectiveness and reliability of their real-time services, such as travel time estimation and route planning.
Despite numerous efforts have been made on congestion prediction, most of them fall short in handling heterogeneous and dynamic spatio-temporal dependencies (e.g., periodic and non-periodic congestions), particularly in the presence of noisy and incomplete traffic data. 
In this paper, we introduce a Congestion Prediction Mixture-of-Experts, \model, to address the above challenges. 
We first propose a sparsely-gated Mixture of Adaptive Graph Learners (MAGLs) with congestion-aware inductive biases to improve the model capacity for efficiently capturing complex spatio-temporal dependencies in varying traffic scenarios. 
Then, we devise two specialized experts to help identify stable trends and periodic patterns within the traffic data, respectively. 
By cascading these experts with MAGLs, \model delivers congestion predictions in a more robust and interpretable manner.
Furthermore, an ordinal regression strategy is adopted to facilitate effective collaboration among diverse experts. 
Extensive experiments on real-world datasets demonstrate the superiority of our proposed method compared with state-of-the-art 
spatio-temporal prediction models. 
More importantly, \model has been deployed in DiDi to improve the accuracy and reliability of the travel time estimation system.
\end{abstract}



\maketitle

\section{Introduction}\label{sec:intro}
Rapid urbanization in recent years has led to an unprecedented increase in traffic volumes, posing significant challenges to modern Intelligent Transportation Systems (ITS). As an integral part of ITS, traffic congestion prediction aims to anticipate future traffic conditions (\ie fast, slow, and congested) on the roads, which plays a pivotal role in human livelihood and urban governance. For example, accurate prediction of traffic congestion enables drivers to make informed trip planning decisions in advance, thereby largely reducing travel time and fuel consumption. Furthermore, it also empowers various decision-making tasks in transportation management, such as route planning~\cite{polestar2020kdd}, 
public transit scheduling~\cite{congtransit2022}, and emergency response planning~\cite{congemergency2015}. As a result, congestion prediction has been extensively studied in both academia and industry.

\eat{
For traffic congestion prediction, its is critical to effectively capture spatio-temporal patterns of traffic data. In the past decade, many efforts have been made to develop advanced Deep Learning~(DL) models for tackling this problem.
To name a few, 
DMLM~\cite{dmlm2021ijcai} formalizes congestion propagation and dissipation process via dynamic attributed graphs and proposes a meta learning enhanced STGNN approach to learn complex spatio-temporal correlations.
DuTraffic~\cite{dutraffic2022cikm} jointly incorporates street view data and STGNN encoded trajectory data to make accurate real-time congestion prediction.
PSE-STGCN~\cite{psestgcn2023sigspatial} models the differences and similarities of periodic congestion patterns and further extracts mid-term and short-term information from congestion event sequences to enhance STGNN.}
Accurate congestion prediction relies on effective modeling of spatio-temporal dependencies within traffic data. In the past decade, many efforts have been made to develop advanced Deep Learning~(DL) models for tackling this problem~\cite{cpsurvey2021jat,cpsurvey2021trpc}. To name a few,
Ma~\etal~\cite{rbmrnn2015plos} leverages deep Restricted Boltzmann Machines~(RBMs) and Recurrent Neural Networks~(RNNs) to model the temporal dynamics of congestion.
Chen~\etal~\cite{pcnn2018tits} incorporates Convolutional Neural Networks~(CNNs) to jointly model the recent and periodic congestion patterns.
Xia~\etal~\cite{dutraffic2022cikm} and Wang~\etal~\cite{sttf2023ijcis} adopt Spatial Temporal Graph Neural Networks~(STGNNs) to capture the intricate traffic propagation patterns for congestion level prediction.

Despite fruitful progress in this field, building industry-level congestion prediction systems still faces the following challenges.
\eat{


}
(1) Urban traffic data exhibit heterogeneous and dynamic spatio-temporal dependencies in non-congestion, periodic congestion and non-periodic congestion scenarios.
\eat{
}
For example, main roads may experience periodic congestion with regular propagation patterns during rush hours, while roads around landmarks (\eg sports stadiums) more often encounter non-periodic congestion, showcasing diverse ranges in space and time depending on specific events.
Even within the same location, the traffic propagation patterns during congested peak hours can be distinct from those in non-congested periods.
Simply introducing a more sophisticated model architecture or increasing the model size is insufficient to handle such dynamic and heterogeneous traffic patterns. 
Besides, a large parameter size also introduces substantial computational costs, which hinders the model deployment in production.
Therefore, it is challenging to develop a cost-effective congestion prediction model for diverse spatio-temporal dependency preservation.
(2) Real-world traffic data suffer from frequent missingness and noises~\cite{urbancomputing2014tist}.
For instance, in ride-hailing platforms like DiDi, real-time traffic features such as congestion levels are calculated from the GPS trajectories of ride-hailing vehicles. 
However, the sparseness of ride-hailing vehicles in specific areas or at particular time slots might lead to data missing issues. Moreover, poor GPS signals and unpredictable driver behaviors can introduce additional noises that could distort the actual traffic patterns.
Therefore, it is challenging to develop congestion prediction models that are robust to these missing values and noises.
(3) Interpretability is critical for the industry-scale deployment of congestion prediction models~\cite{inherentinter2019nature}. 
Stakeholders such as the product manager and customer need to understand why certain areas are predicted to be congested to make informed traffic management or travel decisions.
However, it is difficult for humans to interpret the reasoning process of deep learning models due to their inherent black-box nature.
Therefore, how to enhance the prediction interpretability is another challenge.

To tackle the above challenges, in this paper, we present a Congestion Prediction Mixture-of-Experts~(\model), which consists of three major modules.
First, inspired by the recent success of sparsely-gated Mixture-of-Experts~(MoE) in handling large-scale heterogeneous data~\cite{sparsemoe2017iclr,nlpmoe2022emnlp,cvmoe2021nips}, we propose a Mixture of Adaptive Graph Learners (MAGLs) module. 
By training multiple specialized graph learning experts on varied subgroups of data and selectively activating them on specific samples via a sparse gating mechanism, MAGLs possess a much larger capacity to accommodate heterogeneous and evolving traffic patterns while maintaining superior inference efficiency.
Second, to enhance the model's robustness against potential data missingness and noise, we introduce two specialized experts focusing on capturing stable trends and periodic patterns, respectively. 
By cascading these experts with MAGLs, this approach not only directs MAGL's focus towards more complex samples but also enhances the model's decision-making transparency via the interpretable expert aggregation weights.
\eat{
}
Finally, We adopt ordinal regression strategy~\cite{softord2019cvpr} to alleviate the experts' over-confidence issue caused by their varied inductive biases and the inherent class imbalance, enabling beneficial collaboration among experts.

Our main contributions are summarized as follows.
(1) To the best of our knowledge, this is the first attempt to apply the MoE architecture to an industry-level congestion prediction application.
(2) We propose a progressive expert design grounded in the characteristics of urban traffic data to construct an expressive and scalable congestion prediction model. We further organize the experts cascadingly to boost its robustness and interpretability.
(3) We devise an ordinal regression strategy to balance the confidence of experts in order to prevent the collapse of expert collaboration.
(4) We conduct extensive experiments on real-world traffic datasets to demonstrate the superiority of \model against advanced spatio-temporal prediction models. We further utilize the congestion prediction results to improve the travel time estimation service in production.

\section{Problem Statement}\label{sec:prob_state}
This paper focuses on urban traffic congestion prediction. We first introduce several important definitions as follows. 
\begin{myDef}[Traffic Network]
    The traffic network is defined as a directed weighted graph $\mathcal{G}=(V,E)$, where $v_i\in V$ denotes road link and $e_{ij}\in E$ denotes the adjacent relation between $v_i$ and $v_j$. 
    At time interval $t$, the dynamic traffic features is denoted as $\mathbf{X}^t \in \mathbb{R}^{N \times C}$, where $N=|V|$ and $C$ is the number of dynamic traffic feature type. 
    Besides, we let $\mathbf{X}_i^t \in \mathbb{R}^{C}$ denote the dynamic features of link $v_i.$
\end{myDef}
\begin{myDef}[Congestion Level]
    We assess traffic conditions of links using three discrete congestion levels: fast, slow, and congested, denoted as class 0, 1, and 2, respectively.
\end{myDef} 
With the above concepts, we next formulate the target problem.
\begin{prob}[Congestion Prediction]
Given traffic feature sequence $\mathbf{X}^{t-T_p+1:t}:=( \mathbf{X}^{t-T_p+1}, \mathbf{X}^{t-T_p+2}, ..., \mathbf{X}^t ) \in \mathbb{R}^{T_p \times N \times C}$ from previous $T_p$ time intervals, historical traffic feature set $\mathcal{H}$ and traffic network~$\mathcal{G}$, 
we aim to learn a mapping function $\mathcal{F}(\cdot)$ to predict the congestion level in the future $T_f$ time intervals, 
\begin{align}
    \mathcal{F}: (\mathbf{X}^{t-T_p+1:t}, \mathcal{H};
    \mathcal{G}) \mapsto \hat{\mathbf{Y}}^{t+1:t+T_f} \in \{0,1,2\}^{T_f \times N}.
\end{align}
\end{prob}

\section{Data Description and Analysis}

\begin{figure*}[t] 
\vspace{-3pt}
\centering  
\subfigure[Spatial dependency in peak hours.]{
    \label{subfig:noncong}
    \includegraphics[width=0.23\linewidth]{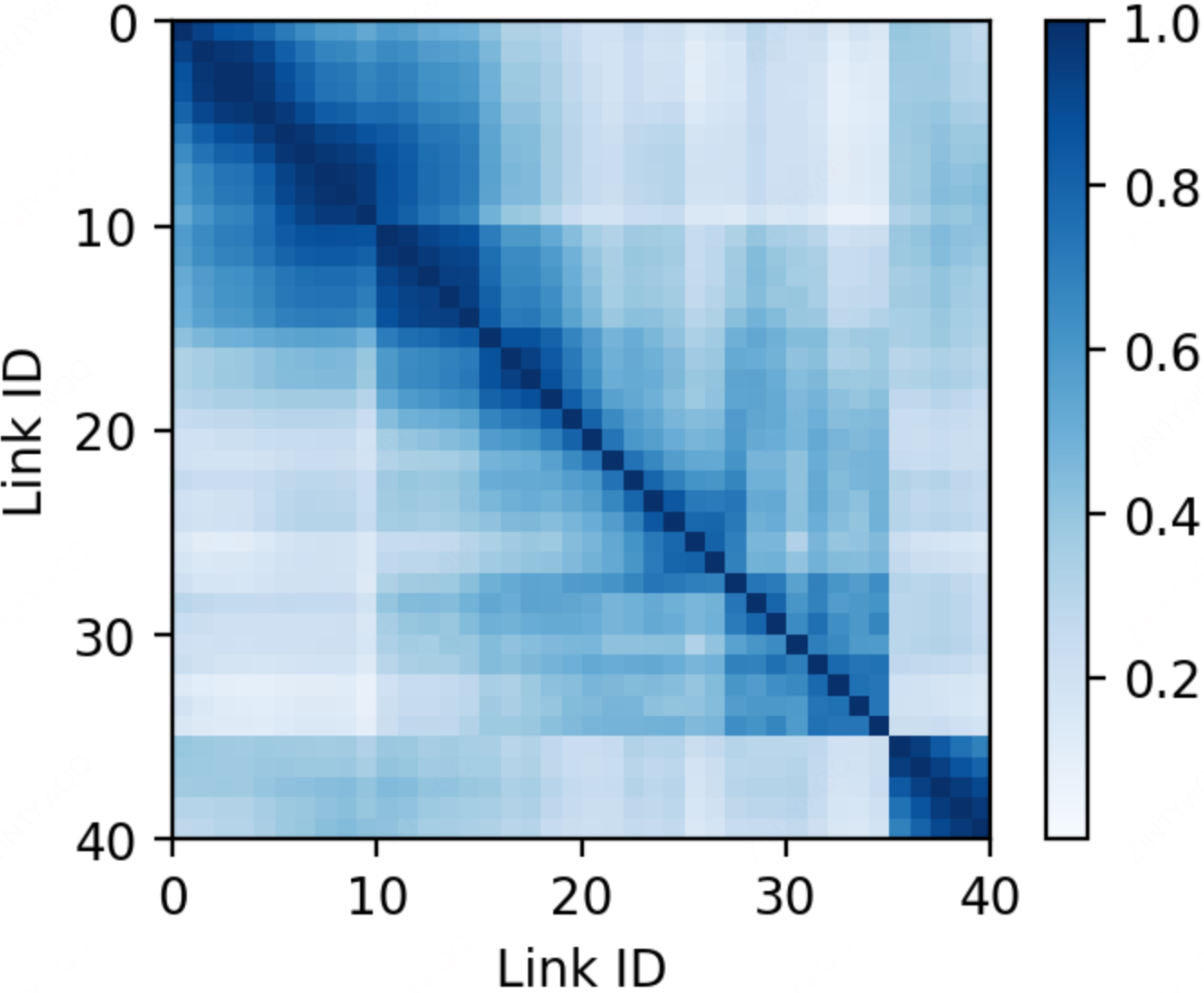}}
\subfigure[Spatial dependency in nonpeak hours.]{
    \label{subfig:cong}
    \includegraphics[width=0.23\linewidth]{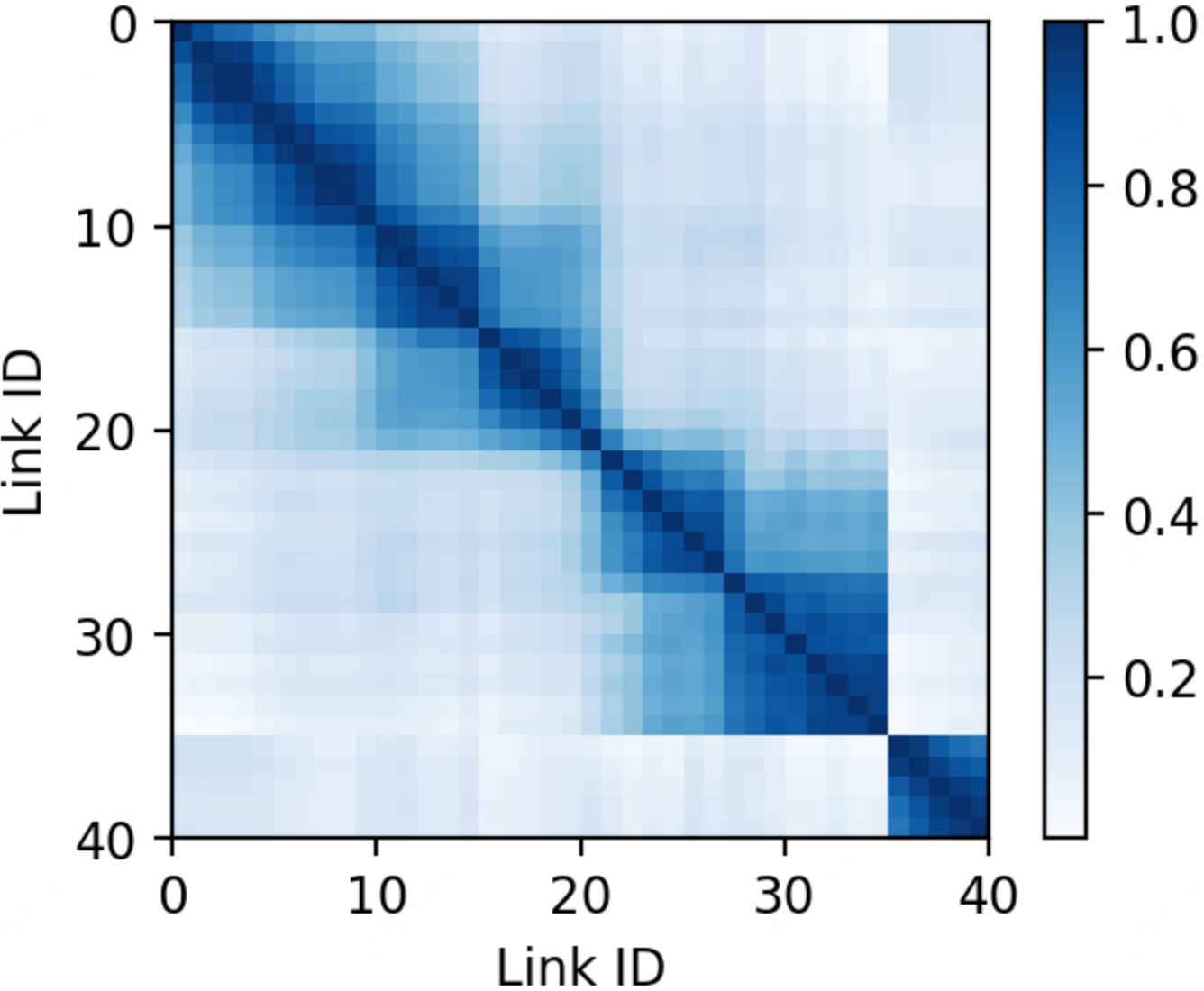}}
\subfigure[Congestion level shift probability.]{
    \label{subfig:cond_shift}
    \includegraphics[width=0.24\linewidth]{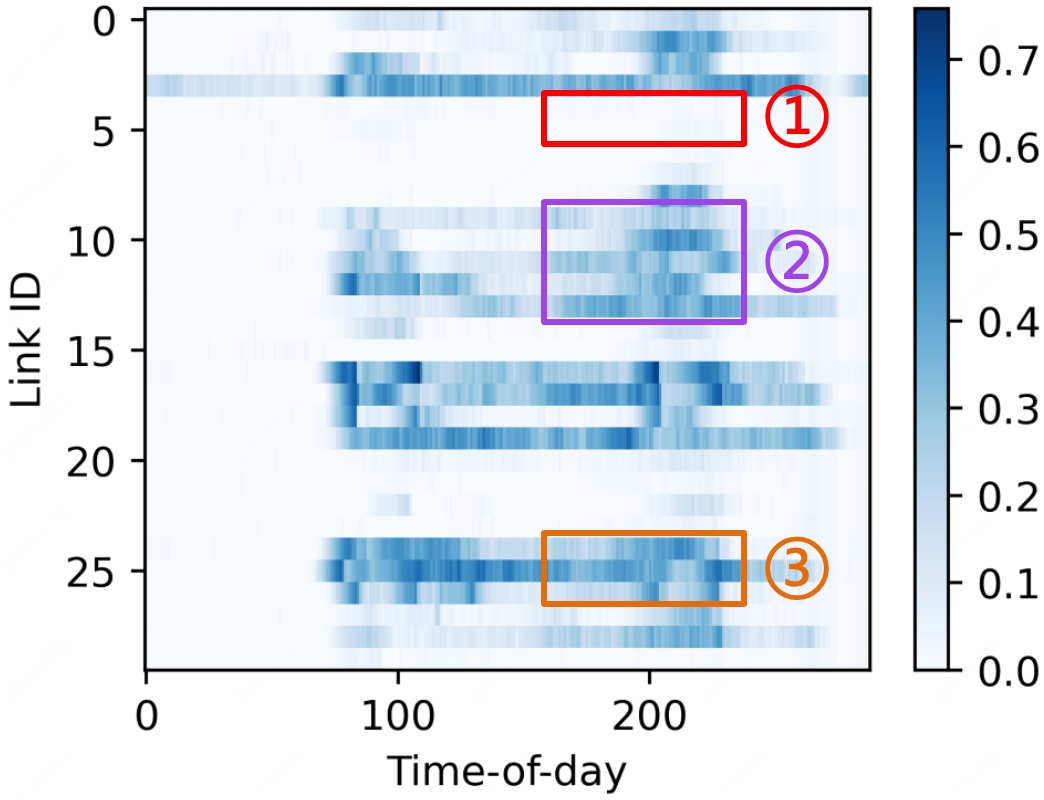}}
\subfigure[Periodicity of congestion.]{
    \label{subfig:cong_prob}
    \includegraphics[width=0.24\linewidth]{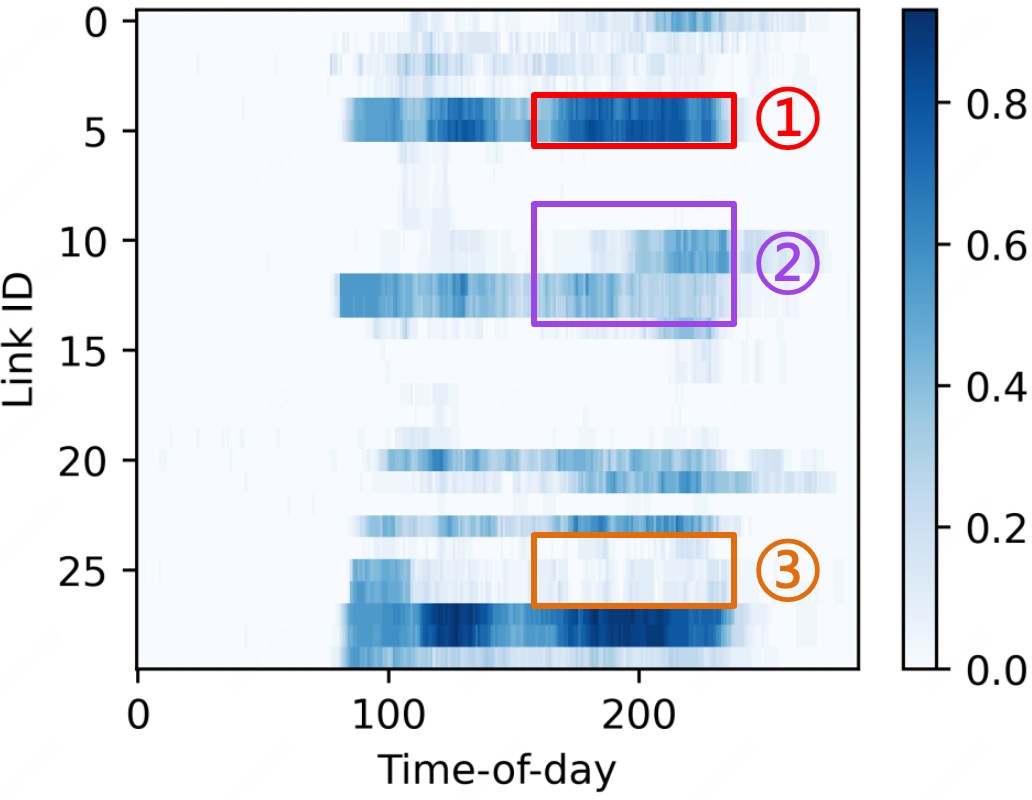}}
    \vspace{-5pt}
\caption{Primary data analysis on Beijing dataset. 
In (a) and (b), deep color indicates higher dependency. In (c), deeper color implies higher instability of traffic condition. In (d), deeper color implies a higher likelihood of periodic congestion. }
\vspace{-5pt}
\end{figure*}

\subsection{Data Description} \label{subsec:data_descript}
We study our problem on real-world datasets collected from Beijing and Shanghai, two metropolises in China. \tabref{tab:stats} summarizes their detailed statistics. The two datasets range from September 24, 2023, to November 03, 2023, and from October 30, 2023, to December 09, 2023, respectively.
They are constructed from trajectory records in DiDi's ride-hailing platform, covering hundreds of links from urban arterial roads, where congestion happens frequently. 

The datasets encompass a range of static link attributes and dynamic traffic features.
The static attributes include link length, width, speed limit, number of lanes, longitude and latitude.
The dynamic traffic features of each link comprise real-time traffic speed and congestion level. 
Traffic speed is computed by averaging the speed of trajectories that traverse a specific road link. 
Congestion level is labeled by practical rules developed by DiDi, taking into account factors like static link attributes and average speed. 
\begin{table}[t]
    \centering
    \caption{Statistics of two real-world datasets.}
    \vspace{-5pt}
    \begin{tabular}{l|c|c}
    \toprule
         & Beijing & Shanghai \\ 
    \midrule 
        \# of Time intervals &  \multicolumn{2}{c}{11808} \\ \hline
        Time interval &   \multicolumn{2}{c}{5 minute}   \\ \hline
        \# of road links &  568  &  707  \\ \hline
        Congestion ratio &  18.2\%  & 6.6\%  \\ \hline
        Missing ratio &  0.7\%  & 2.3\%  \\ 
        \bottomrule
    \end{tabular}
    \vspace{-5pt}
    \label{tab:stats}
\end{table}

\subsection{Feature Construction} \label{subsec:feats}
We use two categories of features for each road link: static features and dynamic features.
The static features comprise link attributes $\mathbf{S} \in \mathbb{R}^{N\times D_s},$ where $D_s$ is the number of static features, and link distance feature $r_{ij}$ is calculated based on the longitude and latitude of link $v_i$ and $v_j.$
The dynamic features are originally updated every 1 minute. We aggregate them into 5-minute intervals by averaging the traffic speed features and selecting the most frequent congestion level.
The recent dynamic features are collected from previous 12 intervals.
The historical features $\mathcal{H}$ are extracted from previous days and weeks, comprising 
$\mathcal{H}_d = \{\mathbf{X}^{(t+1 - d\cdot T_d): (t + T_f - d\cdot T_d)}\}_{1\leq d \leq N_d}$ and $\mathcal{H}_w = \{\mathbf{X}^{(t+1 - w\cdot T_w): (t + T_f - w\cdot T_w)}\}_{1\leq w \leq N_w},$
where $T_d, T_w$ is the number of time intervals in a day and a week, respectively, and $N_d, N_w$ is the maximum number of days or weeks we trace back. In practice, we set $N_d = 4$ and $N_w = 3.$


\subsection{Data Analysis}
In this section, we conduct primary analysis on the Beijing datasets to intuitively illustrate the challenges of congestion prediction.

\subsubsection{Spatio-temporal dependencies} \label{subsubsec:st_depend}
To understand the complex spatio-temporal dependencies in urban traffic data, we sample 40 inter-connected arterial road links that cross multiple city function regions and visualize the Pearson correlation matrix~\cite{pearson2009} of them during peak and non-peak hours, respectively in \figref{subfig:noncong} and \ref{subfig:cong}. 
In both figures, we can observe strong intra-region dependencies (deeper color around the main diagonal) and much weaker inter-region dependencies, which reveals the intricate spatial heterogeneity.
Moreover, by comparing the two figures, we observe a clear dependency variation from peak hours to non-peak hours, 
which motivates us to improve the model capacity for capturing heterogeneous traffic patterns in varied scenarios.

\subsubsection{Stability and periodicity of congestion} \label{subsubsec:stab_per}
\eat{
In \figref{subfig:cond_shift}, we visualize, for 30 sampled links over 288 time intervals of a day, the probability of congestion levels deviating from the current congestion level in the future 1 hour. Deeper
color implies higher instability of traffic condition. 
In \figref{subfig:cong_prob}, we visualize the probability of congestion for the same set of links at different time intervals. Deeper color implies a higher likelihood of periodic congestion.
In both figures, we can observe the high stability and periodicity of traffic condition in late night.
While in the daytime, the stability and periodicity vary significantly. 
Comparison of box 1 in two figures suggests the existence of stationary congestion, where the congested condition persists over a long period.
Comparison of box 2 implies the existence of non-stationary congestion where the traffic condition frequently shifts.
Comparison of box 3 reveals the existence of non-stationary evolution in non-congested scenarios. 
Moreover, the three boxes in \figref{subfig:cong_prob} also indicates the co-existence of periodic and non-periodic congestion. These findings necessitates an adaptive utilization of the trend and periodic information to strength the model performance.}
In \figref{subfig:cond_shift} and \ref{subfig:cong_prob}, we analyze over 30 links the probability of congestion level shift and congestion occurrence across 288 daily time intervals, respectively. Deeper colors indicate higher traffic instability or higher likelihood of periodic congestion, respectively. 
We can observe the stability and periodicity of traffic at night, contrasting with the significant variability observed during daytime. 
Specifically, Box 1 comparison indicates the existence of persistent stationary congestion, while Box 2 comparison suggests non-stationary congestion with frequently shifting traffic conditions. Box 3 comparison reveals the non-stationary evolution in non-congested scenarios. 
Moreover, the three boxes in \figref{subfig:cong_prob} also reflect the existence of both periodic and non-periodic congestion. 
These findings necessitates an adaptive utilization of trend and periodic patterns to enhance model accuracy without affecting its learning on complex scenarios.

\section{The Proposed Approach}\label{sec:approach}

\subsection{Model Overview}
\begin{figure*}[t] 
\vspace{-5pt}
\centering  
\subfigure[Overall framework of \model.]{
    \label{fig:overall_framework}
    \includegraphics[width=0.59\linewidth]{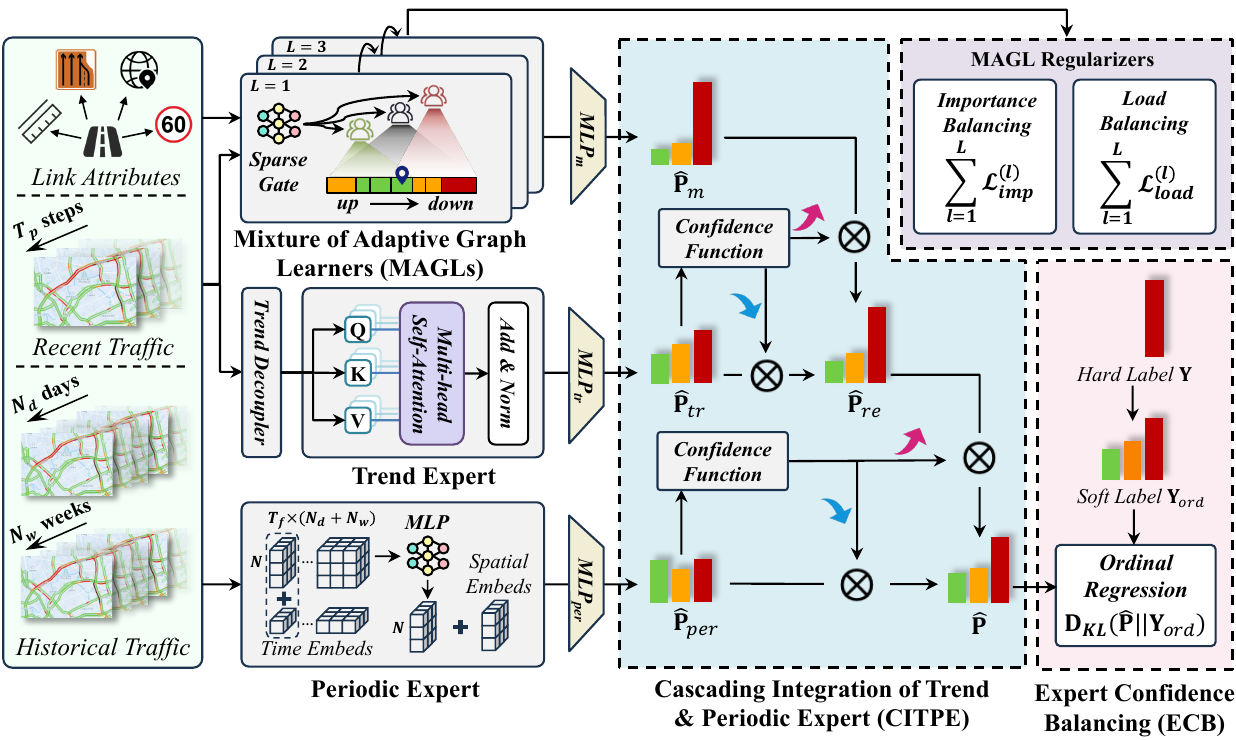}}
\quad
\subfigure[Overview of a MAGL layer.]{
    \label{fig:MAGL_layer}
    \includegraphics[width=0.35\linewidth]{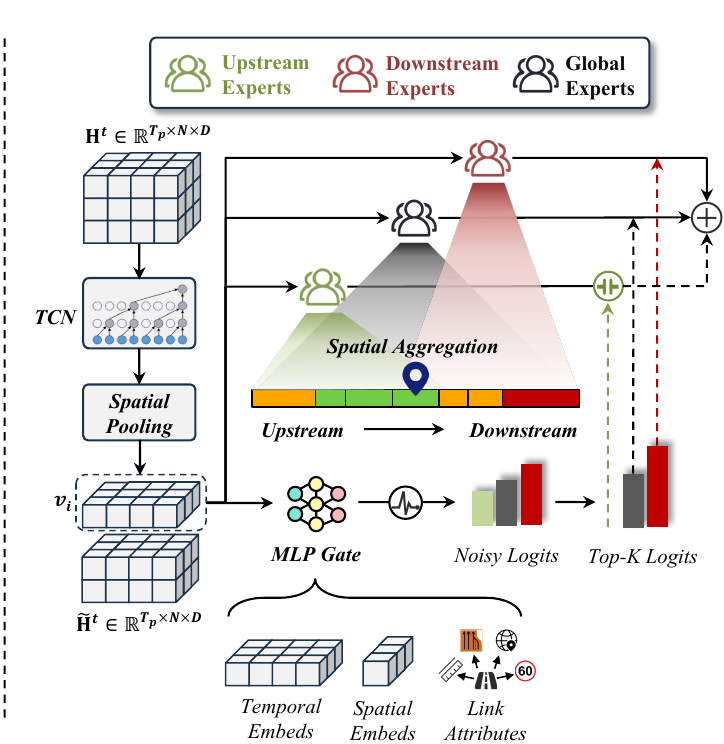}}
    \vspace{-5pt}
\caption{Overall framework of CP-MoE.}
\vspace{-5pt}
\end{figure*}

\figref{fig:overall_framework} depicts the overall framework of \model, 
which consists of three major modules: 
(1) \emph{Mixture of Adaptive Graph Learners (MAGLs)}:
It comprises multiple MAGL layers built upon the sparsely-gated MoE architecture that selectively route samples to specialized graph learning experts for comprehensive and efficient exploration of spatio-temporal dependencies.
(2) \emph{Cascading Integration of Trend and Periodic Experts (CITPE)}:
It adaptively integrates two specialized experts to capture trend and periodic patterns, enhancing the model's robustness to deal with corrupted data.
(3) \emph{Expert Confidence Balancing~(ECB)}: It harnesses ordinal regression to guide the experts in recognizing ordinal relations among congestion levels, mitigating overconfidence in their predictions and fostering effective expert collaboration.

\eat{
\figref{fig:overall_framework} depicts the overall framework of \model, which consists of a mixture of strength-varying experts that are aggregated in a cascading manner.
First, we design a periodicity modeling module, which is composed of a periodicity router and periodicity expert. The router is responsible for determining whether the input sample can be simply predicted by the expert based only on periodic experience. 
If the periodicity is not sufficient, our second module, mixture of trend and event experts, will be activated to further explore the high-order spatio-temporal dependencies for prediction. The trend expert encodes low frequency signals of the target link for prediction, while multiple event experts with varied inductive biases are activated when neighboring high frequency signals are significant for modeling future variation of congestion levels.
Finally, we adopt two confidence-aware strategies to ensure effective collaboration among different experts. 
}

\subsection{Mixture of Adaptive Graph Learners}
\eat{This module comprises a stack of MAGL layers to capture complex high-order spatio-temporal dependencies.
\figref{fig:MAGL_layer} presents an overall structure of each MAGL layer, which consists of a learnable sparse gate that selectively activate various types of graph learning experts based on fine-grained spatio-temporal contextual inputs to extract high-level spatio-temporal patterns.
We first formally define the MAGL layer in \secref{subsubsec:MAGL_overview}, then discuss the details of gate inputs and expert designs in \secref{subsubsec:gate_input} and \secref{subsubsec:graph_expert}.}
\eat{
In each layer, we first employ a gated Temporal Convolutional Network~(TCN) to encode recent traffic sequences. Then, these encodings along with learnable spatial and temporal embeddings are fed into a sparse gating mechanism to adaptively activate suitable graph learners for spatial pattern extraction. Moreover, we infuse distinct inductive biases into the experts to boost the learning of gating strategy.
}
As shown in \figref{fig:MAGL_layer}, a MAGL layer leverages a learnable sparse gate to select specific experts from a shared expert pool for each link at a specific time slot. Formally, a MAGL layer is defined as
\begin{align}
    {\mathbf{H}_i^t}^{(l+1)} = \sum_{n=1}^{N_e} G_n \left( I\left({\mathbf{H}_i^t}^{(l)} \right) \right) \cdot E_n\left({\mathbf{H}_i^t}^{(l)} \right),
\end{align}
where $N_e$ is the number of experts, $G(\cdot)$ denotes the sparse gate function and $G_n(\cdot)$ is the $n$-th element of the output vector from $G(\cdot)$, which determines the importance of the $n$-th expert $E_n(\cdot)$; $I(\cdot)$ indicates a profiling function designed to augment the context features provided to the gate; ${\mathbf{H}_i^t}^{(l)} \in \mathbb{R}^{T_p \times D}$ is the output of the $l$-th layer for link $v_i$ at time interval $t$, and ${\mathbf{H}_i^t}^{(0)} = \operatorname{FC}(\mathbf{X}_i^{t-T_p+1:t})$, where $\operatorname{FC}(\cdot)$ stands for a fully connected layer.

In practice, we stack $L$ MAGL layers and utilize a Multi-Layer Perceptron~(MLP) to generate the congestion level logits 
for all road links over next $T_f$ time steps,
\begin{align}
    \hat{\mathbf{P}}_{m}^{t+1:t+T_f} = \operatorname{MLP}_m({\mathbf{H}_i^t}^{(L)}).
\end{align}
We detail the gate function and expert design below. For simplicity, we omit the layer index in the superscript.

\subsubsection{Sparse gate with fine-grained inputs}  \label{subsubsec:gate_input}
The sparse gate function in MAGL layer is crucial for learning diverse spatio-temporal patterns present in traffic data. A good gate function should route input samples to the most suitable experts under specific context~\cite{moesurvey2022}. To achieve this goal, we curate a collection of fine-grained context features as gate inputs to enhance the sample distinguishability.

As convolution networks is sensitive to high-frequency signals (\eg unexpected congestion)~\cite{vitwork2022iclr}, we first leverage gated Temporal Convolution Networks~(TCN)~\cite{gwnet2019ijcai} to extract temporal dynamics representations ${\mathbf{H}_i^t}'$ from the input $\mathbf{H}_i^t$.
The details of gated TCN are presented in~\appref{app:tcn}. Afterwards, we derive the short-term spatio-temporal context by using a lightweight sum operator $\tilde{\mathbf{H}}_i^t = \sum_{v_j \in \mathcal{N}_k} {\mathbf{H}_j^t}'$,
where $\mathcal{N}^k_i$ denotes the $k$-hop neighbors of link $v_i$. In practice, we find that the sum operator is more efficient and effective than other learnable aggregator functions, \eg GNNs, the empirical evidence of which can be found in \appref{app:sum_operator}.

However, short-term information may lack reliability and distinguishability across various contexts~\cite{stnorm2021kdd,stid2022cikm}. Consequently, we further incorporate three types of stable features: (1) static link attributes $\mathbf{S}$, (2) a learnable spatial embedding $\mathbf{E}^s \in \mathbb{R}^{N \times D_l}$ that encapsulates unique spatial characteristics, and (3) Time-of-Day~(ToD) and Day-of-Week~(DoW) embeddings $\mathbf{E}^{ToD}\in \mathbb{R}^{288 \times D_l}, \mathbf{E}^{DoW} \in \mathbb{R}^{7 \times D_l}$ that encode regular temporal patterns.
Overall, the input of the gate function \wrt link $v_i$ at time $t$ can be written as
\begin{align}
    I(\mathbf{H}_i^t) = \tilde{\mathbf{H}}_i^t \  \| \ \operatorname{MLP}_s(\mathbf{S}_i) \ \| \ \mathbf{E}^s_i \ \| \ \mathbf{E}^{ToD}_{t} \ \| \ \mathbf{E}^{DoW}_{t}. \label{eqn:gate_input}
\end{align}

Based on the resulting features $\mathbf{c}_i^t=I(\mathbf{H}_i^t)$, we follow previous work on MoE~\cite{sparsemoe2017iclr} and apply noisy top-K gating mechanism, \ie
\begin{gather}
    G(\mathbf{c}_i^t) = \operatorname{Softmax} (\operatorname{TopK}(\operatorname{MLP}_g(\mathbf{c}_i^t) + \epsilon \cdot \operatorname{Softplus}(\operatorname{MLP}_n(\mathbf{c}_i^t)) ),
\end{gather}
where $\operatorname{Softplus}(\cdot)$ is an activation function~\cite{softplus2011aistat}, the output logits of $\operatorname{MLP}_g(\cdot)$ is added with Gaussian noise $\epsilon \in \mathcal{N}(0, 1)$ to avoid model collapse (\ie over-reliance on a few experts), and $\operatorname{TopK}(\cdot)$ sparsely activates $K$ experts based on the largest entries in the noisy logits.

\subsubsection{Congestion-aware graph experts} \label{subsubsec:graph_expert}
As discussed in \secref{subsubsec:st_depend}, there exists a substantial discrepancy in spatio-temporal patterns between congestion and non-congestion scenarios. Such pattern discrepancy may introduce noises or even mutually contradictory knowledge, making it difficult to train a unified model that perfectly recognizes patterns across different scenarios. To tackle the challenge, we devise three groups of graph learning experts, each endowed with a dedicated inductive bias that enables specialization in a particular pattern type. 

Specifically, our expert design is motivated by the following key insight: traffic congestion usually spreads from downstream to upstream links, whereas traffic flow freely propagates from upstream to downstream links during non-congestion periods. Based on this insight, we assign two specialized expert groups, namely \emph{upstream experts} and \emph{downstream experts}, to model the two distinct propagation dynamics.
In general, each expert $E(\cdot)$ is implemented with an edge-aware graph attention network~\cite{gat2018iclr,revisit_hgnn2021kdd}, which adaptively aggregate neighboring information into the target link via
\begin{gather}
    E(\mathbf{H}^t_i) = \sum_{j \in \mathcal{N}_i} \alpha_{ij} \mathbf{W}_j \mathbf{H}^t_j + \mathbf{H}^t_i, \label{eqn:gat} \\
    \alpha_{i j}=\frac{\exp \left(\operatorname{LeakyReLU}\left( \mathbf{a}^T\left[\mathbf{W} \mathbf{H}^t_i\left\|\mathbf{W} \mathbf{H}^t_j\right\| \mathbf{W}_r r_{ij}\right]\right)\right)}{\sum_{k \in \mathcal{N}_i} \exp \left(\operatorname{LeakyReLU}\left(\mathbf{a}^T\left[\mathbf{W} \mathbf{H}^t_i\left\|\mathbf{W} \mathbf{H}^t_k\right\| \mathbf{W}_r r_{ik}\right]\right)\right)},
\end{gather}
where $\mathbf{W}$, $\mathbf{W}_r$ and $\mathbf{a}^T$ are learnable mappings. For upstream experts, $\mathcal{N}_i$ consists of the upstream links of $v_i$, while for downstream experts, $\mathcal{N}_i$ only encompasses the downstream links.

However, the graph topology built from the road network is often noisy and incomplete, which may not reflect the actual relationships among road links. Therefore, we additionally assign a group of \emph{global experts} to specialize in identifying latent propagation patterns~\cite{gwnet2019ijcai,agcrn2020nips}.
Each global expert is equipped with a unique learnable link embedding $\mathbf{E}^s \in \mathbb{R}^{N\times D_l}$ to encode the inherent spatial characteristics. 
The hidden dependency between link $v_i$ and $v_j$ can then be inferred via
\begin{align}
    \alpha_{ij} = \operatorname{Softmax}(\operatorname{ReLU}(\mathbf{E}_i^s {\mathbf{E}^s_j}^T)),
\end{align}
which can be used for spatial aggregation by following~\equref{eqn:gat}.

\subsection{Cascading Integration of Trend and Periodic Experts}
\eat{
}
\eat{Despite MAGLs' strengths, their effectiveness diminishes when short-term traffic propagation patterns are blurred by data missingness and noises. Instead, identifying stable trend patterns of the target link becomes more beneficial as it is unaffected by neighboring interference.
Moreover, when there exists severer short-term data anomalies, focusing on periodic patterns derived from historical traffic data emerges as the most viable solution.}
Despite MAGLs' strengths, their effectiveness diminishes when short-term traffic propagation patterns are distorted by noises and missing data. In such cases, the trend and periodic patterns are more useful for forecasting, as they are insensitive to neighboring interference.
Driven by this insight, we improve model robustness by constructing a \emph{trend expert} and a \emph{periodic expert} to capture stable trend and periodicity, respectively.
On the other hand, the analysis in \secref{subsubsec:stab_per} indicates the existence of intricate non-stationary and non-periodic traffic patterns, requiring greater efforts from MAGL for accurate prediction. This necessitates an adaptive approach to cascade the trend and periodic expert with MAGL. We introduce the detailed design as follows.

\subsubsection{Trend decoupling and modeling}
The trend of traffic condition is represented by low-frequency signals within the short-term traffic observations, which is often coupled with high-frequency signals that fluctuate over time.
Therefore, we first adopt Discrete Wavelet Transform~(DWT)~\cite{wavelet1989siam} to decompose input $\mathbf{X}^{t+T_p-1 : t}$ into components at varied frequency scales, and use inverse DWT to reconstruct the trend signals $\mathbf{R}^{t+T_p-1 : t}$ from low-frequency components.
We defer details of this procedure to \appref{app:dwt}.
After that, we adopt a Multi-head Self-Attention~(MSA)~\cite{transformer2017nips} based trend expert followed by a MLP layer to output the prediction logits of future traffic conditions, 
\begin{align}
\hat{\mathbf{P}}_{tr}^{t+1:t+T_f} = \operatorname{MLP}_{tr}(\operatorname{MSA}(\mathbf{R}^{t+T_p-1 : t})).
\end{align}


\subsubsection{Periodicity modeling}
\eat{
As introduced in \secref{subsec:feats}, $\mathcal{H}$ contains the historical traffics from the closest days and weeks, which can also reflect the recent traffic context, \eg{a large raining persisting for several days}.
Therefore, we adopt MLPs along with learnable spatial and temporal embeddings to more thoroughly extract periodic information and produce the predictions logits as 
\begin{align}
    \hat{\mathbf{P}}_{per}^{t+1:t+T_f} = \operatorname{MLP}_{per}(\mathcal{H},\mathbf{E}^s,\mathbf{E}^{ToD},\mathbf{E}^{DoW}).
\end{align}
Please refer to \appref{app:per_expert} for more details of periodic expert.
}
When there exists severer corruption within short-term data, the underlying periodic patterns of historical data facilitate a more robust prediction.
The historical traffic features $\mathcal{H}$, as introduced in~\secref{subsec:feats}, encompass global periodic patterns driven by daily human routines like morning commutes and local patterns influenced by external factors, such as recent weather variations.
To capture such multifaceted periodicity, we incorporate learnable spatio-temporal embeddings and develop an efficient MLP-based periodic expert for future prediction,
\begin{align}
    \hat{\mathbf{P}}_{per}^{t+1:t+T_f} = \operatorname{MLP}_{per}(\mathcal{H},\mathbf{E}^s,\mathbf{E}^{ToD},\mathbf{E}^{DoW}).
\end{align}
Please refer to \appref{app:per_expert} for more implementation details.

\subsubsection{Cascading expert integration}\label{subsubsec:conf_adjust}
The primary goal of adaptively integrating trend and periodic expert is to ensure that these experts dominate corrupted data, while MAGL are reserved for complex patterns.
However, the distribution similarity between the two types of data make it difficult to learn such an ideal routing strategy without explicit supervision signals.
Motivated by the recent study~\cite{weakstrongmoe2023}, we mitigate this issue by leveraging the experts' prediction confidences to determine their influence on the final prediction.

Concretely, the trend or periodic expert will be assigned a weight computed from its output logits via a learnable score function $C(\hat{\mathbf{P}}) = \operatorname{MLP}_c ( D(\hat{\mathbf{P}}) ).$
Here $D(\cdot)$ is a dispersion function that calculates the variance and negative entropy of logits to measure prediction confidence.
$\operatorname{MLP}_c(\cdot)$ is trained to map the dispersion to an expert weight within the range $[0, 1].$
The time indices of logits $\hat{\mathbf{P}}$ are omitted for brevity.
Furthermore, the ordering of expert aggregation is guided by two principles:
(1) Activate stronger experts only when the confidence levels of all weaker experts are low in order to promote focused learning of complex patterns; (2) The periodic expert is considered weaker than the trend expert due to its inaccessibility to the latest traffic observations.
These principles lead to a cascading expert aggregation strategy, which derives the final prediction logits of \model from the outputs of different experts as
\begin{align}
    \hat{\mathbf{P}}_{re} &= C_2(\hat{\mathbf{P}}_{tr})\hat{\mathbf{P}}_{tr} + \left(1 - C_2(\hat{\mathbf{P}}_{tr})\right)\hat{\mathbf{P}}_m, \\
    \hat{\mathbf{P}} &= C_1(\hat{\mathbf{P}}_{per})\hat{\mathbf{P}}_{per} + \left(1 - C_1(\hat{\mathbf{P}}_{per})\right) \hat{\mathbf{P}}_{re},
\end{align}
where $C_1(\cdot)$ and $C_2(\cdot)$ are two learnable confidence functions.
Notably, this approach also possesses great interpretability as the model decision process can be explained by the weights of experts.

\eat{As trend or periodic patterns are more resilient to missingness and noises, the two corresponding experts will take over easy-to-learn but imperfect samples, which in turn helps stabilize MAGLs' optimization process for more complex samples.}

\subsection{Ordinal Regression for Expert Confidence Balancing}\label{subsec:ord}
\eat{
The experts may exhibit varying confidence levels due to the discrepancies in their assigned data subgroups and architectural inductive biases.
In addition, the inherent data imbalance between congestion and non-congestion class may exacerbate the risk of over-confidence~\cite{overconf2021cvpr}.
Consequently, the outputs of overconfident experts could disproportionately influence the final output.}

\eat{The effectiveness of CITPE may be impacted by imbalances in experts' confidence levels.
These imbalances can stem from the experts' distinct inductive biases and differing data imbalance between congestion and non-congestion classes in their assigned data subgroups~\cite{overconf2021cvpr}. 
Consequently, overconfident experts might diminish others' contributions, leading to skewed final outputs.}
The diversified inductive biases among experts, coupled with the varying degree of congested class imbalance within their assigned data subsets, can lead to considerable disparities in confidence levels. Consequently, overly confident experts may undermine the contributions of others, potentially leading the CITPE module to make biased predictions.



To this end, we leverage ordinal regression~\cite{softord2019cvpr} to mitigate the overconfidence issue of experts.
This approach smooths one-hot labels into soft labels by redistributing a portion of probability from target class to other classes, thereby reducing experts' confidence in a single class. 
Moreover, the classes closer to the target class will be assigned a larger probability to preserve the natural ordering among classes. Such a strategy further enriches each class in the label space with additional information from nearby classes~\cite{imbreg2021icml,ranksim2022icml}, effectively reducing overconfidence caused by class imbalance.

Formally, the $i$-th element of the one-hot label encoding is adjusted to $y_{ord}[i] = e^{-\phi(i,y)} / \sum_{j} e^{-\phi(j,y)},$
where $y$ denotes the target class and $\phi(\cdot,\cdot)$ is a pre-defined distance metric that penalizes the probability of distant classes.
In the context of congestion prediction, given the finite number of classes, $\phi(\cdot,\cdot)$ can be determined through hyperparameter tuning.
In practice, we further constrain the distance metric to satisfy $\phi(i,j) + \phi(j,k) = \phi(i, k), \ \forall 0 \le i \le j \le k,$ thereby narrowing the tuning space to $\{\phi(i,i+1)\}_{i \ge 0}$.


\begin{table*}[t] \footnotesize 
\caption{The 12-step congestion prediction performance on Beijing and Shanghai datasets. The best results are in bold, the second-best are underlined, and the third-best are marked with a star.}
\vspace{-5pt}
\begin{tabular}{c|ccccc|ccccc}
\toprule
& \multicolumn{5}{c|}{Beijing} & \multicolumn{5}{c}{Shanghai} \\ \cline{2-11} 
\multirow{-2}{*}{Model} & Accuracy(\%) & Recall(\%) & Precision(\%) & W-F1(\%) & C-F1(\%) & Accuracy(\%) & Recall(\%) & Precision(\%) & W-F1(\%) & C-F1(\%)  \\ 
\midrule

{CT} & 80.25 & 70.79 & 70.67 & 69.44 & 70.73 & 87.59 & 59.05 & 58.85 & 65.20 & 58.95 \\ 
{HA} & 78.57 & 55.52 & 73.10 & 62.65 & 63.11 & 88.25 & 53.54 & 61.01 & 64.41 & 57.04 \\ 
\midrule
DCRNN & 81.27$\pm$0.10 & 71.78$\pm$0.95 & 71.46$\pm$0.74 & 69.32$\pm$0.31 & 71.61$\pm$0.15 & 88.67$\pm$0.07 & 52.24$\pm$1.68 & 67.41$\pm$1.16 & 65.27$\pm$0.43 & 58.83$\pm$0.70 \\ 
ASTGNN & 81.53$\pm$0.27 & 72.14$\pm$3.29 & 72.05$\pm$2.74 & 70.33$\pm$0.74 & 71.97$\pm$0.52 & 88.83$\pm$0.65 & 65.13$\pm$3.73* & 62.04$\pm$4.23 & 68.19$\pm$1.08 & 63.31$\pm$1.11 \\ 
GWNet & 84.13$\pm$0.16 & 76.10$\pm$2.16 & 76.21$\pm$1.25 & 74.26$\pm$0.40 & 76.12$\pm$0.50 & 90.81$\pm$0.05 & 63.55$\pm$2.85 & 73.03$\pm$2.33* & 72.55$\pm$0.56 & 67.86$\pm$0.70 \\ 
AGCRN & 84.30$\pm$0.11 & 77.77$\pm$1.33* & 75.76$\pm$1.10 & 74.48$\pm$0.24 & 76.73$\pm$0.20 & 90.72$\pm$0.09 & 64.51$\pm$1.18 & 71.71$\pm$1.16 & 72.54$\pm$0.27 & 67.90$\pm$0.27 \\ 
STID & 83.81$\pm$0.16 & 75.42$\pm$0.99 & 75.49$\pm$0.79 & 73.60$\pm$0.36 & 75.45$\pm$0.30 & 90.95$\pm$0.06* & 63.02$\pm$1.35 & \underline{73.70$\pm$1.06} & 72.75$\pm$0.27* & 67.92$\pm$0.39 \\ 
STWave & 84.21$\pm$0.13 & 77.36$\pm$1.18 & 75.58$\pm$1.05 & 74.38$\pm$0.18 & 76.44$\pm$0.15 & 90.56$\pm$0.07 & 64.01$\pm$1.64 & 71.10$\pm$1.42 & 72.05$\pm$0.30 & 67.34$\pm$0.37 \\ 
ST-MoE & 84.39$\pm$0.10* & \underline{77.92$\pm$1.12} & 75.88$\pm$0.91 & 74.60$\pm$0.22* & 76.89$\pm$0.18* & 90.83$\pm$0.11 & 64.60$\pm$1.01 & 71.98$\pm$1.12 & 72.70$\pm$0.29 & 68.09$\pm$0.27* \\
STAEformer & \underline{84.71$\pm$0.14} & 77.73$\pm$1.68 & \underline{76.85$\pm$1.39} & \underline{75.21$\pm$0.25} & \underline{77.26$\pm$0.30} & \underline{91.05$\pm$0.07} & \underline{65.49$\pm$1.69} & 72.26$\pm$1.55 & \underline{73.38$\pm$0.15} & \underline{68.67$\pm$0.26} \\
\midrule
DuTraffic & 81.66$\pm$0.15 & 72.58$\pm$1.10 & 71.76$\pm$0.70 & 69.41$\pm$0.25 & 72.17$\pm$0.21 & 88.62$\pm$0.10 & 52.54$\pm$1.46 & 66.70$\pm$1.36 & 65.02$\pm$0.40 & 58.78$\pm$0.70 \\
STTF & 84.19$\pm$0.16 & 76.40$\pm$2.01 & 76.28$\pm$1.34* & 74.30$\pm$0.40 & 76.34$\pm$0.58 & 90.64$\pm$0.09 & 63.12$\pm$2.67 & 72.87$\pm$2.19 & 72.35$\pm$0.57 & 67.65$\pm$0.93 \\

\midrule
\textbf{CP-MoE} & \textbf{85.09±0.13} & \textbf{80.30±1.01} & \textbf{76.82±0.76} & \textbf{75.92±0.38} & \textbf{78.52±0.20} & \textbf{91.20±0.12} & \textbf{66.38±1.39} & \textbf{73.90±1.01} & \textbf{74.09±0.61} & \textbf{69.94±0.42} \\
\bottomrule
\end{tabular}
\label{table:overall_performance}
\vspace{-3pt}
\end{table*}

\subsection{Optimization Objectives}
The optimization objectives of \model consist of two parts.
The first part is the ordinal regression loss to encourage a balanced confidence levels among experts, which computes the Kullback-Leibler~(KL) divergence between the \model's output logits $\hat{\mathbf{P}}$ and the ordinally smoothed congestion level labels $\mathbf{Y}_{ord},$ \ie
\begin{align}
    \mathcal{L}_{ord} = D_{KL}(\hat{\mathbf{P}} \| \mathbf{Y}_{ord}).
\end{align}
The second part comprises two types of expert balancing regularizers to prevent the model collapse issue of MAGLs~\cite{sparsemoe2017iclr}. To be more specific, each MAGL layer is equipped with an \emph{importance balancing} loss $\mathcal{L}_{imp},$ which limits the variation in the weights assigned to different graph experts, and a \emph{load balancing} loss $\mathcal{L}_{load},$ which ensures equitable activation frequencies across experts. Formally, 
\begin{align}
    \mathcal{L}_{imp} = CV_{j} (\sum_{x\in \mathcal{B}}{G_j(x)}), \ 
    \mathcal{L}_{load} = CV_{j} (\sum_{x\in \mathcal{B}}{Pr_j}),
    \label{eqn:moe_load_loss}
\end{align}
where $CV(\cdot)$ is the coefficient of variation and $Pr_j$ is the probability of the $j$-th expert been activated over a batch of samples $\mathcal{B}$.
More details on the differentiability of $\mathcal{L}_{load}$ can be found in Appendix A of work~\cite{sparsemoe2017iclr}.

Overall, we train \model by jointly optimizing the objectives
\begin{align}
    \mathcal{L} = \mathcal{L}_{ord} + \lambda_1 \sum_{l=1}^L\mathcal{L}_{imp}^{(l)} + \lambda_2 \sum_{l=1}^L\mathcal{L}_{load}^{(l)},
\end{align}
where $\mathcal{L}_{imp}^{(l)}$ and $\mathcal{L}_{load}^{(l)}$ are the importance balancing loss and load balancing loss of the $l$-th MAGL layer,
$\lambda_1$ and $\lambda_2$ are hyperparameters that controls the extent of expert balancing.


\section{Experiments}\label{sec:exp}
We conduct comprehensive experiments to answer the following research questions.
\textbf{RQ 1}: How is the overall performance of \model compared with the SOTA baselines on real-world datasets?
\textbf{RQ 2}: How robust is \model \wrt{varied missing and noise ratios?} 
\textbf{RQ 3}: How do different modules in \model affect the model performance?
\textbf{RQ 4}: How is \model's interpretability?
\textbf{RQ 5}: Can \model benefit travel time estimation in the ride-hailing service?

\subsection{Experimental Setup}

\subsubsection{Metrics} \label{subsubsec:metrics}
For the congestion prediction task, we use Accuracy, Recall, Precision, F1-score~(C-F1) and Weighted F1-score~(W-F1) for evaluation.
Specifically, Recall, Precision, C-F1 are calculated \wrt congested class. W-F1 is the average of the F1-scores for three classes, with the weights being 0.2, 0.2 and 0.6, respectively.
For the travel time estimation task, we use Mean Average Error~(MAE), Root Mean Squared Error~(RMSE), 
and Bad Case Rate~(BCR)~\cite{probtte2023kdd}.

\subsubsection{Baselines} 
We compare our proposed framework with the following 
baselines.
\textbf{Rule-based strategies:} (1) CurrentTime~(CT): It uses the present congestion level as the predictions for future $T_f$ time intervals; (2) HistoricalAverage~(HA): It predicts the congestion level of link $v_i$ at time $t$ as its most frequent historical congestion level at time interval $t$ in the training set.
\textbf{General STGNNs: }
DCRNN~\cite{dcrnn2018iclr}, ASTGNN~\cite{astgnn2022tkde}, GWNet~\cite{gwnet2019ijcai}, AGCRN~\cite{agcrn2020nips}, STID~\cite{stid2022cikm}, STWave~\cite{stwave2023tkde}, ST-MoE~\cite{stmoe2023cikm} and STAEformer~\cite{staeformer2023cikm}. 
\textbf{Congestion prediction methods: }
DuTraffic~\cite{dutraffic2022cikm} and STTF~\cite{sttf2023ijcis}.
Please refer to \appref{app:baseline} for more details on these methods and \appref{app:imple} for implementation details of \model.


\subsection{Overall Performance Comparison (RQ 1)}

Table \ref{table:overall_performance} reports the 12-step prediction results of \model and all the compared baselines \wrt five evaluation metrics.
Since the performance is very close in the offline and online environments, we only report the offline results.
From \tabref{table:overall_performance}, we can make the following observations.
First, the proposed model outperforms all baselines in terms of all metrics on two datasets.
Significant improvements in Recall and C-F1 highlight its adeptness at learning complex traffic congestion patterns, while enhancements in Precision and W-F1 confirm its ability to maintain accuracy across congested and non-congested scenarios, owing to the tailored sparse gating mechanism and specialized expert designs.
Besides, the methods which aggregate spatial information on pre-defined graph topology~(DCRNN, ASTGNN) perform significantly worse than those on learnable adaptive graphs~(GWNet, AGCRN, STWave), underscoring the complexity of congestion propagation patterns and the need for enlarging the capacity of spatial modeling module.
With a more capable architecture, STAEformer and ST-MoE achieves further improvements on both datasets. However, due to the lack of task-customization, they still underperform \model.
In addition, on the Shanghai dataset with a higher data missing ratio, STID outperforms complex adaptive graph based methods. This is because STID's learnable embeddings can capture regular spatio-temporal patterns resilient to data anomalies.
Our method shows further improvement, which validates the efficacy of CITPE module in enhancing the model robustness without compromising the ability to learn complex patterns.
Moreover, the two congestion prediction models underperform the selected STGNNs, indicating their limited customization and the universality of state-of-the-art STGNNs.

\subsection{Robustness Check (RQ 2)}
To further justify \model's robustness during training, we synthesize various levels of data missingness or noises by randomly masking or flipping $p\%$ observed traffic features in the training set of the Beijing dataset, where $p \in \{20, 40, 60, 80\}$.
The performance variations of representative models are shown in \figref{fig:robust}. Due to similar variation trends \wrt C-F1 and W-F1, we visualize only the C-F1 performance.
Clearly, all methods experience performance degradation as $p$ increases, as they encounter more unrealistic traffic patterns.
AGCRN struggles with high noise or missing data due to its over reliance on complex spatial dependency modeling. Removing CITPE module from \model~(CP-MoE-\emph{WoC}) performs well at low levels of missingness or noise, but fails significantly under extreme data noise, mirroring AGCRN's limitation. 
Differently, ASTGNN and STID capture corruption-resilient long-term patterns by explicitly modeling periodic features or incorporating learnable spatio-temporal embeddings, achieving robust performance across varied settings.
\model achieves comparable performance fluctuation with STID in missing data scenarios, and significantly less fluctuation in noisy conditions compared to all baselines. Meanwhile, it consistently outperforms baselines in all scenarios.
These results affirm the value of trend and periodicity modeling for robustness and emphasize the efficacy of adaptive expert aggregation in maintaining ability to learn complex patterns.

\begin{figure}[t]
\centering  
\includegraphics[width=0.90\linewidth]{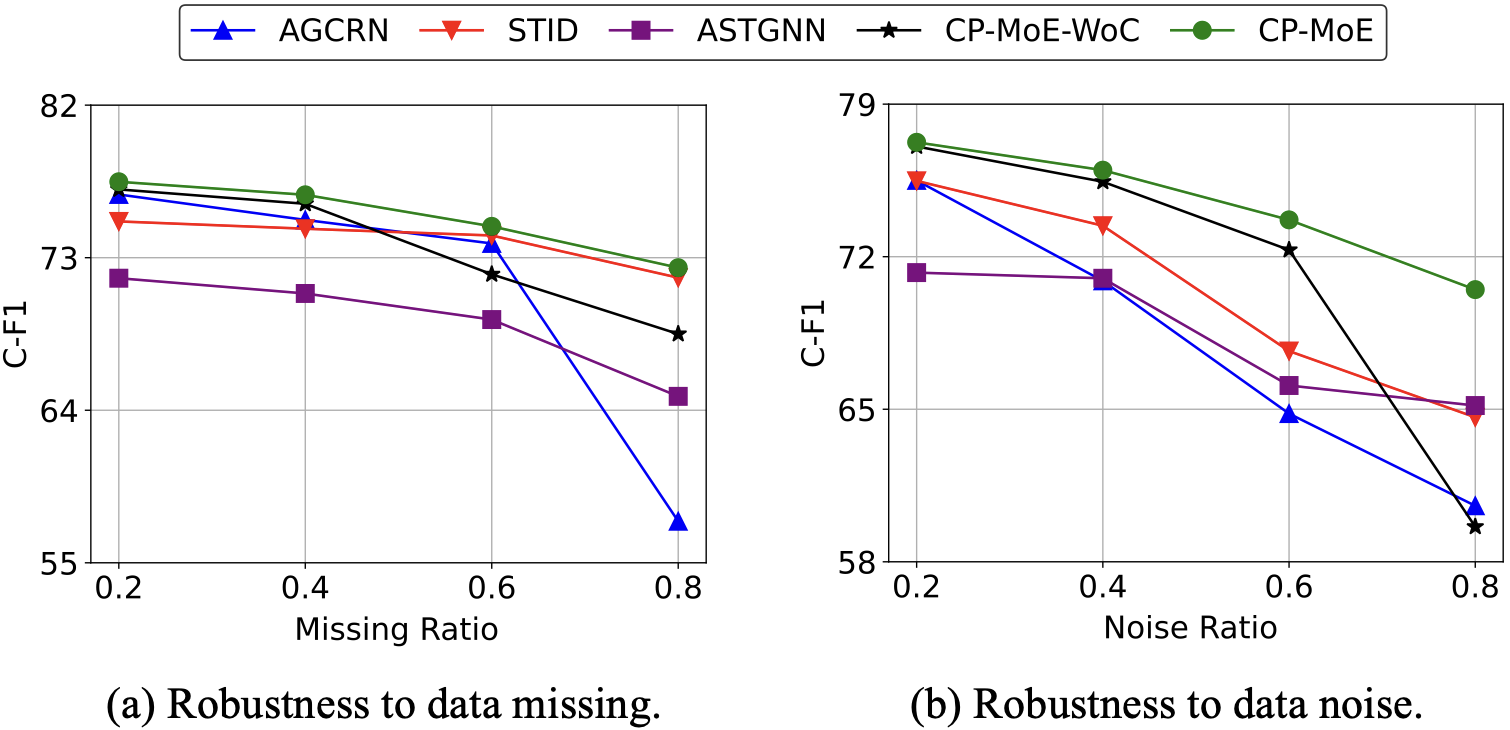}
\vspace{-5pt}
\caption{Robustness check on the Beijing dataset.}
\label{fig:robust}
\end{figure}

\begin{figure}[t]
\centering  
\includegraphics[width=0.98\linewidth]{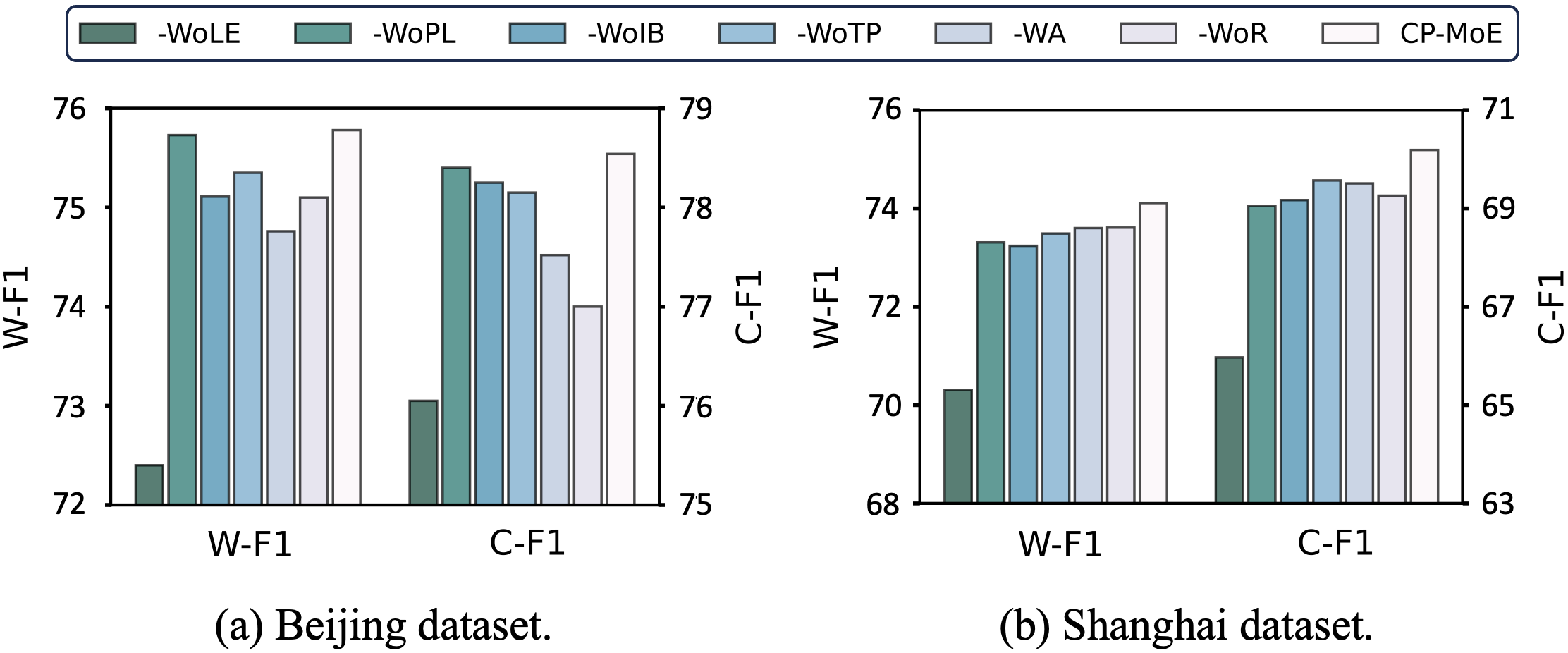}
\vspace{-5pt}
\caption{Ablation study on Beijing and Shanghai datasets.}
\vspace{-5pt}
\label{fig:ablation}
\end{figure}

\eat{
\begin{figure}[t]
\centering  
\subfigure[]{
    \label{subfig:missing}
    \includegraphics[width=0.45\linewidth]{figures/robust/missing.pdf}}
\quad
\subfigure[]{
    \label{subfig:noise}
    \includegraphics[width=0.45\linewidth]{figures/robust/noise.pdf}}
\caption{Robustness check on the Beijing dataset.}
\label{fig:robust}
\end{figure}
}

\subsection{Ablation Study (RQ 3)}
To validate the effectiveness of each module in \model, we compare the performances of the following variants:
(1)~\emph{-WoLE} removes both spatial and temporal learnable embeddings in MAGLs' gate inputs.
(2)~\emph{-WoPL} removes the local pooling when compiling MAGL's gate inputs.
(3)~\emph{-WoIB} removes the spatial inductive biases in MAGL's experts.
(4)~\emph{-WoC} removes the CITPE module.
(5)~\emph{-WA} replaces the confidence-based expert aggregation with simple average aggregation. 
(6)~\emph{-WoR} removes the ordinal regression.
\eat{
\begin{enumerate}
    \item \model\emph{-WoPL} removes the local pooling when compiling the inputs of MAGL's gate.
    \item \model\emph{-WoLE} removes both spatial and temporal learnable embeddings when compiling the gating inputs.
    \item \model\emph{-WoIB} removes the spatial inductive biases in MAGL.
    \item \model\emph{-WA} replaces the confidence-based expert aggregation with simple average aggregation. 
    \item \model\emph{-WoC} removes the trend and periodic experts.

    \item \model\emph{-WoR} removes the ordinal regression.
\end{enumerate}
}
As shown in \figref{fig:ablation}, we can make the following observations.

\subsubsection{The efficacy of the MAGL design.}
First, removing the learnable embeddings and spatial pooling results in performance degradation, confirming the importance of discriminative inputs for effective gating strategy learning. 
Additionally, removing expert inductive biases in MAGLs leads to performance drops, underscoring the value of expert specialization in handling diverse traffic conditions.

\subsubsection{The efficacy of the CITPE design.}
First, the performance drop caused by removing CITPE module validates the importance of capturing stable trend and periodic patterns to model robustness. 
Second, replacing the confidence-based aggregation with average aggregation also causes the performance drop, which verifies the importance of adaptive expert integration for the model to excel in both data-imperfect and complex scenarios.

\subsubsection{The efficacy of ordinal regression.}
We also observe the improvement achieved by the ordinal regression strategy. This validates its effectiveness in alleviating the over-confidence issue of experts and fostering collaborative efforts among.

\eat{
\begin{figure}[t]
\centering  
\subfigure[Results on Beijing dataset.]{
    \label{subfig:ablation_bj}
    \includegraphics[width=0.45\linewidth]{figures/ablation/beijing.pdf}}
\quad
\subfigure[Results on Shanghai dataset]{
    \label{subfig:ablation_sh}
    \includegraphics[width=0.45\linewidth]{figures/ablation/shanghai.pdf}}
\caption{Ablation study on the two datasets.}
\label{fig:robust}
\end{figure}
}

\subsection{Interpretability Analysis (RQ 4)}
In this part, we provide both global and local analysis of experts weights to justify the inherent interpretability of \model.

\begin{figure}[t] 
\begin{minipage}{1.0\linewidth}
\centering  
\subfigure[Trend expert dominated sample distribution.]{
    \label{subfig:trend_dist}
    \includegraphics[width=0.33\textwidth]{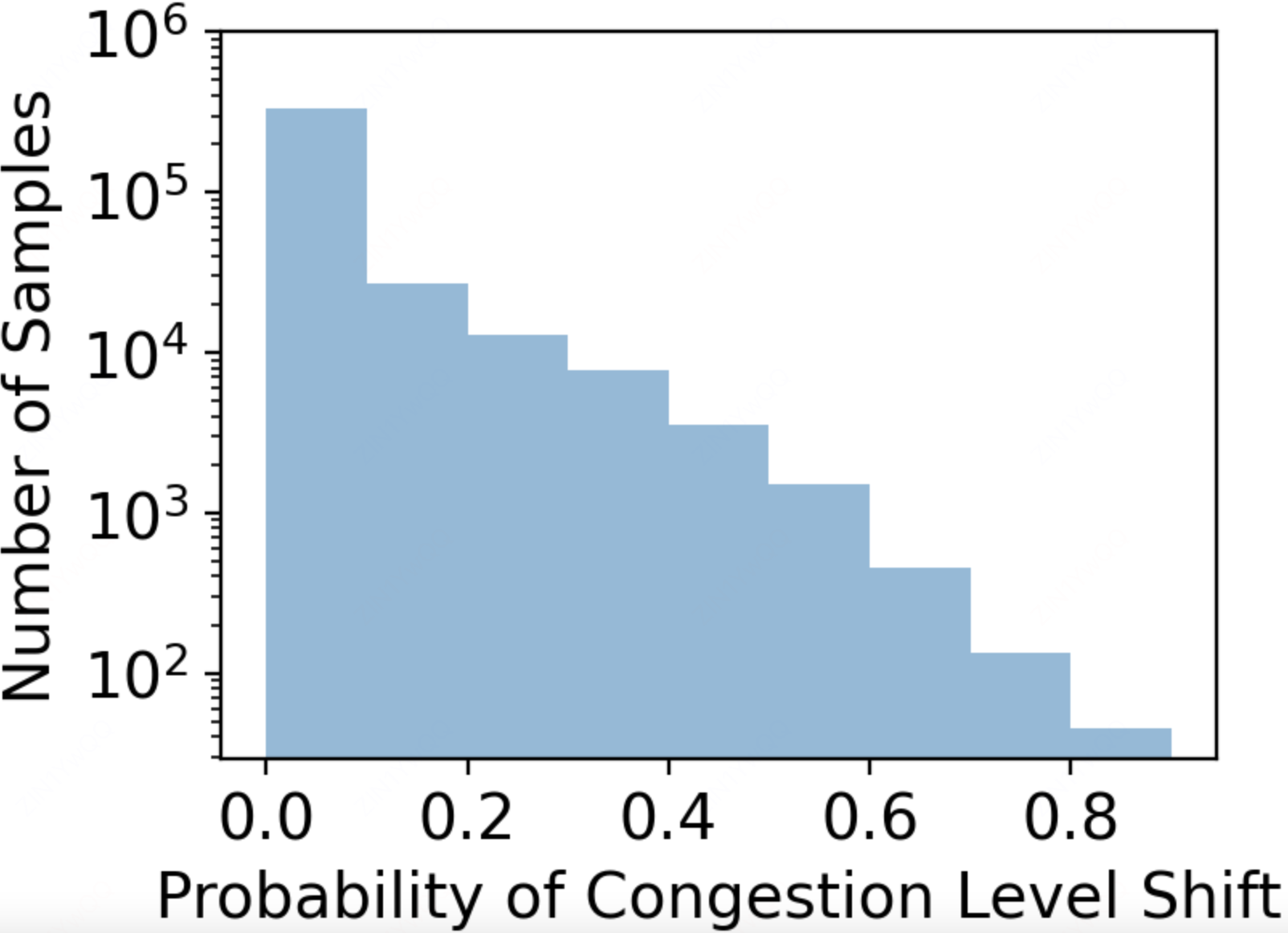}}
\subfigure[Periodic expert dominated sample distribution.]{
    \label{subfig:per_dist}
    \includegraphics[width=0.31\textwidth]{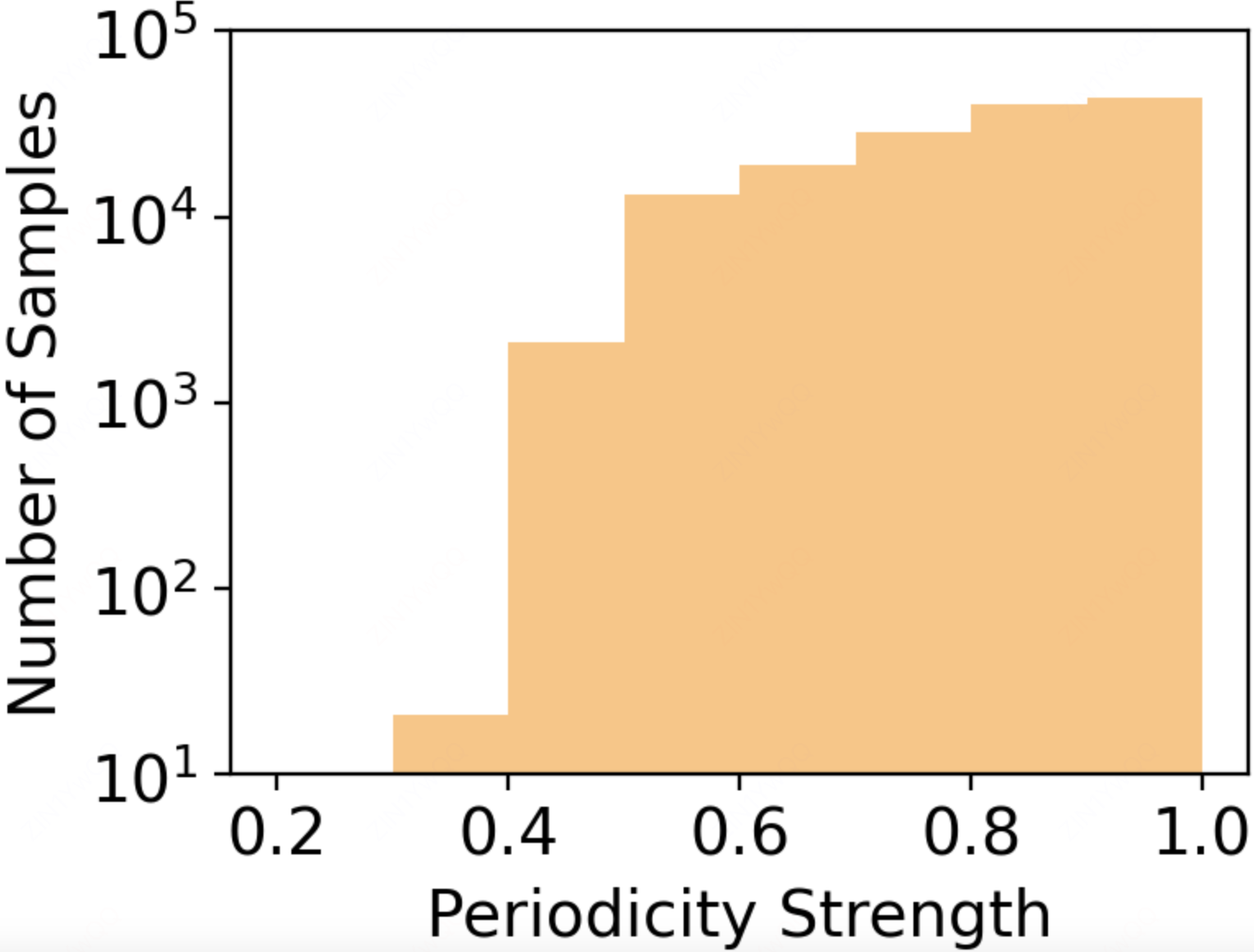}} 
\subfigure[Trend and periodic experts' weight distributions.]{
    \label{subfig:trend_per_gates}
    \includegraphics[width=0.31\textwidth]{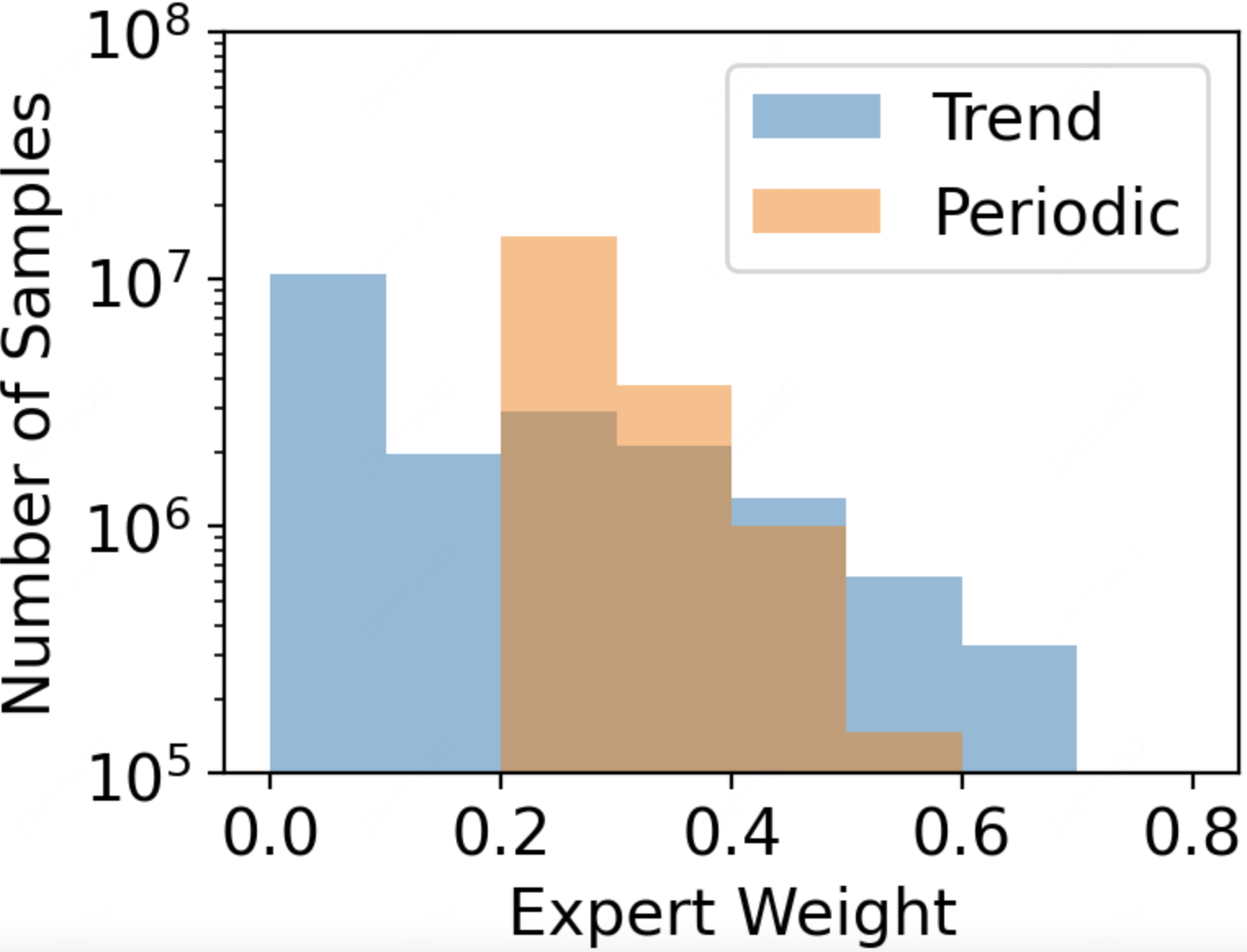}}
\end{minipage}
\vspace{-5pt}
\caption{Expert weight distribution on Shanghai Dataset.}
\label{fig:weight_dist}
\vspace{-3pt}
\end{figure}

\begin{figure}[t] 
\centering  
\includegraphics[width=0.48\textwidth]{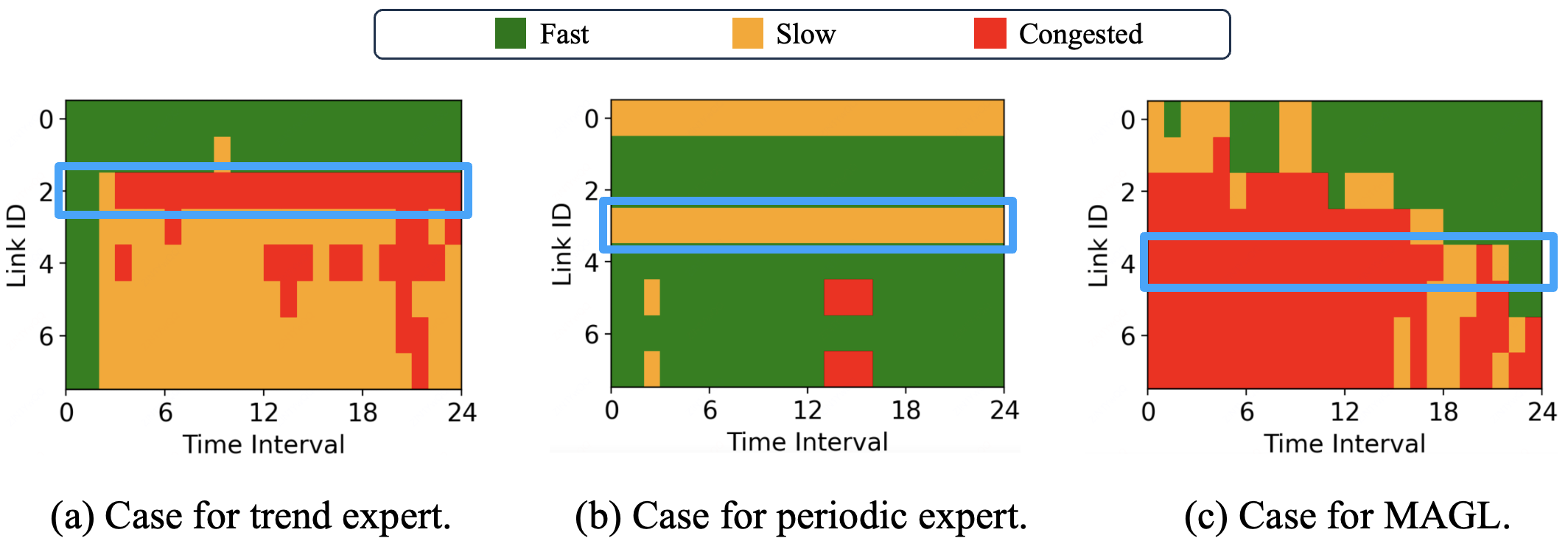}
\caption{Case study on Shanghai Dataset. Each case shows the traffic evolution of 8 links that are arranged sequentially from upstream to downstream. 
CP-MoE predicts the congestion levels of the link marked in blue box for the future 12-24 intervals, based on the observations in previous 12 intervals.}
\label{fig:case_study}
\end{figure}

\subsubsection{Expert weight distribution}
\eat{
We visualize the weight distribution of trend and periodic experts on the testing set in \figref{subfig:trend_per_gates}.
Each of the two experts dominates the prediction for a few cases, with its weight large than 0.5. 
Specifically, the cases dominated by trend expert are undergoing stable traffic evolution, while those dominated by periodic expert show strong periodic patterns, as shown in \figref{subfig:trend_dist} and \ref{subfig:per_dist}.
Moreover, in most cases, the weight of trend or periodic expert is much lower than MAGLs, which accords with the stronger capacity of MAGLs.
These alignments between \model's confidence-based expert weights and human's decision logic~(\ie selecting the specialized expert for handling specific tasks) validates \model's inherent interpretability.} 
By examining the weights of trend and periodic experts, we discover that both experts dominate the prediction for a few samples, with their weights larger than 0.5. 
Specifically, as shown in \figref{subfig:trend_dist} and \ref{subfig:per_dist}, the samples dominated by trend expert are mostly undergoing stable traffic evolution, while those dominated by periodic expert show strong periodic patterns.
On the other hand, \figref{subfig:trend_per_gates} reveals the predominant role of MAGLs on a majority of samples.
These observations confirm that the expert weights generated by the model offer insightful and human-understandable interpretations for its decisions, thereby validating the inherent interpretability of \model.


\subsubsection{Local interpretation study}
In \figref{fig:case_study}, we present three cases where \model's predictions are dominated by the trend expert, periodic expert, and MAGLs respectively, as reflected by the expert weights. 
\eat{
Specifically, \figref{fig:utility}(a) presents a case where the target link experiences prolonged congestion, while its adjacent links remain fast or slow-moving. In such situation, the spatial information aggregation by MAGL introduces noise, whereas the trend expert, relying solely on the target link's stable trend, can achieve accurate predictions. 
Differently, in \figref{fig:utility}(b), despite the stable trend, the trend expert may fail to predict confidently due to the rarity of prolonged slow-moving states. In contrast, the periodic expert is more likely to predict it correctly as we find that the target link regularly experience long-lasting slow-moving patterns at this time of a day. 
In \figref{fig:utility}(c), MAGLs can effectively capture the clear propagation patterns between upstream and downstream links for accurate prediction. In contrast, the trend and periodic experts struggle to predict future traffic variations 
solely based on the target link's past states. 
These case studies further demonstrate the efficacy of the expert weights in providing practical and meaningful interpretations of model decisions.
}
In Figure \ref{fig:case_study}(a), the target link undergoes prolonged congestion while its adjacent links vary significantly. MAGLs are not suitable for this case since over spatial aggregation may introduce noise, whereas the trend expert, focusing on the target's stable trend, can predict accurately.  
Figure \ref{fig:case_study}(b) shows a rare prolonged slow-moving case which the trend expert struggle. But the periodic expert can predict accurately due to the pattern's high periodicity strength discovered from the link's historical.
In Figure \ref{fig:case_study}(c), MAGLs can capture clear traffic propagation patterns between links for accurate predictions, while trend and periodic experts might fail by solely relying on target link histories. 
These cases verify the efficacy of expert weights for meaningful model interpretations.

\subsection{Online Travel Time Estimation Test (RQ 5)}
We further conduct online experiments to justify the utility of \model's prediction results on Travel Time Estimation~(TTE), one of the fundamental services in DiDi that directly influences various downstream applications and user experiences.
DiDi's current TTE model utilizes a self-attention mechanism to process static and real-time traffic attributes, including congestion levels, at the start of the order for end-to-end TTE~\cite{ieta2023kdd,probtte2023kdd}. 
Despite DiDi's TTE systems being highly optimized, accurately predicting travel time for longer orders, particularly in non-periodic congestion scenarios, remains a system bottleneck. Our study aims to address this gap by enhancing the TTE system's ability to anticipate future road conditions.

\begin{table}[t]\small
    \centering
    \caption{Utility of CP-MoE on travel time estimation.}
    \vspace{-5pt}
    \begin{tabular}{l|l|ccc}
    \toprule
        Criterion & Model & MAE(sec) & RMSE(sec) & BCR(\%) \\ 
    \hline \hline
        \multirow{2}{*}{$ATT\in[40,50]$} & Base & 222.33 & 295.73 & 7.10 \\ 
        ~ & Base-\emph{WCP} & \textbf{221.87} & \textbf{294.61} & \textbf{7.04} \\ \hline
        \multirow{2}{*}{$ATT\in[50,60]$} & Base & 290.58 & 385.52 & 8.59 \\ 
        ~ & Base-\emph{WCP} & \textbf{288.20} & \textbf{381.56} & \textbf{8.52} \\ \hline
        \multirow{2}{*}{$ATT\in[60,+\infty]$} & Base & 463.23 & 631.10 & 13.48 \\ 
        ~ & Base-\emph{WCP} & \textbf{454.39} & \textbf{617.06} & \textbf{13.12} \\
    \hline \hline
        \multirow{2}{*}{$DR\in[0.5,1.0]$} & Base & 236.94 & 338.24 & 8.93 \\ 
        ~ & Base-\emph{WCP} & \textbf{229.71} & \textbf{327.13} & \textbf{8.37} \\
    \bottomrule
    \end{tabular}
    \vspace{-5pt}
    \label{tab:utility}
\end{table}

\subsubsection{Congestion prediction integration strategy.}
We propose to expose the TTE model to potential future traffic variations by replacing the original congestion level features with \model's predicted ones.
\eat{
}
Specifically, we begin by grouping the orders in the validation set $\mathcal{O}$ based on their estimated travel times $\hat{t}_e$, which are calculated using the historical average speed of the road links and the route length. Each order group $i$ corresponds to the travel time interval $[5i \text{ min}, 5(i+1) \text{ min})$, where $0 \leq i \leq 12$. For orders falling into group $i$, we replace with \model's congestion predictions on their relevant links at future time step $t$, where $0 \leq t \leq 12$, and identify the optimal replacement time $t_i$ that minimizes the MAE of the TTE model. During testing, an order is first matched to its group $i$ based on its $\hat{t}_e$. Then, the congestion predictions at time $t_i$ are fed into the TTE model for prediction.
Notably, training an industry-level TTE model from scratch requires a huge effort. Our strategy is much easier to integrate and apply at scale.

\subsubsection{System Deployment}
We utilized Spark~\cite{zaharia2016apache} to develop an efficient data processing pipeline with two principal steps: (i) extraction of link-level traffic features from ride-hailing car trajectories at one-minute intervals, and (ii) compilation of diverse traffic data every five minutes for future congestion prediction via \model. The deployed CP-MoE conducts prediction on circular road links within Beijing's 5th Ring Road, covering areas with over 500,000 daily ride-hailing orders. These forecasts are generated for the next hour (12 time steps) and are updated every five minutes using data from the preceding hour.
\model and the optimal replacement time of congestion level features in the TTE model are daily updated. Once the updation is completed, the model is pushed to online servers to provide real-time congestion predictions. For each TTE query, its congestion level features are now selected from CP-MoE's predictions. For the complete online processing pipeline of the TTE system, please refer to Appendix A.4.3 in our previous work~\cite{ieta2023kdd}.

\subsubsection{Online testing results.}
In \tabref{tab:utility}, we report the TTE performances of the base TTE model and our strategy~(-\emph{WCP}) on one week's ride-hailing orders with Actual Travel Time~(ATT) exceeding 40 minutes. Our strategy achieves consistent improvements over the base model across all metrics.
\eat{
}
We further select the orders with more than half of the links experiencing a congestion level deviation from departure to arrival, which we denote as order set $\{DR\in [0.5,1]\}$ where $DR$ stands for Deviation Ratio. 
Our strategy achieves more substantial improvements on them, verifying the efficacy of exposing future traffic conditions  to the TTE model, particularly when these conditions exhibit significant variations.
More importantly, the effectiveness of the TTE enhancement strategy in turn corroborates the utility of \model in predicting future traffic conditions in real-world production environment.

\section{Related Work}\label{sec:related}
\eat{
\TODO{(1)too long. (2)is spatiotemporal GNN relevant to your work? sounds MoE is more relevant.}
\wz{(1) Shortened; (2) As our focus is improving congestion prediction rather than providing an improved version of MoEs that possessed generality, we might not have to list MoE related works. STGNN is relevant as we leverage some spatio-temporal modeling techniques, and also compare with STGNN baselines.}}
\textbf{Traffic congestion prediction}.
Existing research on congestion prediction spans various formulations, such as post-congestion propagation prediction~\cite{dmlm2021ijcai}, congestion event prediction~\cite{congevent2023aaai}, and congestion level prediction~\cite{dutraffic2022cikm}. In this paper, we focus on congestion level prediction.
Early studies leverage data mining techniques and shallow machine learning models to capture traffic patterns and uncertainties, including pattern mining~\cite{freqpat_cong2016itsc}, clustering~\cite{clustercong2016}, hidden Markov models~\cite{hmmcong2014trpc} and Bayesian networks~\cite{bayescong2016trr}. While fast to implement and interpretable, they fail to capture the nonlinear spatio-temporal dependencies.
Deep Learning (DL) models have spurred numerous studies to enhance congestion prediction accuracy~\cite{cpsurvey2021trpc}.
Classical DL models such as RNNs and CNNs have been applied to capture both short-term and long-term temporal dynamics~\cite{rbmrnn2015plos,pcnn2018tits}.
Recent DL methods focus on capturing intricate spatio-temporal propagation patterns.
Cheng~\etal~\cite{deeptrans2018ijcnn} combines CNNs with RNNs to extract the spatial-temporal features, while Di~\etal~\cite{cpm2019mdm} further enhances CNNs with mined congestion propagation matrix.
Xia~\etal~\cite{dutraffic2022cikm} incorporates Graph Convolutional Networks~(GCNs), RNNs and MSA to model trajectory data.
Li~\etal~\cite{psestgcn2023sigspatial} adopts TCNs, GCNs and learnable spatio-temporal embeddings to capture short-term and long-term patterns.
Wang~\etal~\cite{sttf2023ijcis} uses Transformer to directly handle 3D spatio-temporal feature tensors. 
However, training a single model is limited in capturing diverse traffic patterns and keeping robustness to data anomalies. The black-box nature of DL models also limits their practical application.

\textbf{Spatio-temporal graph neural networks}.
STGNNs adeptly capture the intricate spatio-temporal data dependencies by integrating graph learning and temporal learning methods, facilitating advanced analytics and forecasting in various applications.
In terms of spatial modeling, existing research can be categorized into static graph based and adaptive graph based methods.
Static graph based methods aggregate information from spatial neighbors determined by pre-defined graphs. For instance, 
DCRNN~\cite{dcrnn2018iclr} conducts graph diffusion enhanced RNN on the traffic network structure for traffic prediction.
ASTGNN~\cite{astgnn2022tkde} performs graph convolution on the traffic network weighted by the dynamic node feature similarity.
Adaptive graph based methods automatically discover hidden dependencies among graph nodes from data. For example,
GWNet~\cite{gwnet2019ijcai}, AGCRN~\cite{agcrn2020nips} and MTGNN~\cite{mtgnn2020kdd} generate adaptive adjacency matrix through the similarities among learnable node embeddings.
MugRep~\cite{mugrep2021kdd} conducts hierarchical graph learning to capture multi-level urban spatio-temporal dynamics.
BigST~\cite{bigst2024vldb} further boosts the adaptive graph learning to linear complexity for large-scale applications.
STWave~\cite{stwave2023tkde} proposes a disentanglement module to enhance the spatio-temporal modeling.
ST-MoE~\cite{stmoe2023cikm} integrates multiple STGNNs into an MoE framework, which only takes observed traffic features as gate inputs and does not support flexible MoE integration.
Some recent works revisit the necessity of GNN in STGNN architecture and instead use MLP~\cite{stid2022cikm} or Transformer~\cite{staeformer2023cikm} to conduct spatio-temporal modeling.
The robustness risk of STGNN training is also revealed in work~\cite{ASTFA2022nips}.
In this work, we elevate the model capacity with tailored sparse-gated spatio-temporal MoE layers. Moreover, compared with general STGNNs, we incorporate task-customized expert designs to boost the utility, robustness and interpretability of congestion prediction.
\eat{
\subsection{Mixture-of-Experts}
ST-MoE \cite{stmoe2023cikm} first incorporates the MoE framework into traffic prediction.

\subsection{Spatio-temporal OOD Generalization}
STNSCM \cite{STNSCM2023aaai}
CauSTG \cite{caustg2023kdd} is most similar with our work, which learns local and global invariant relations by ensembling models trained on hierarchical data environments. 
However, (i) the model ensemble inevitably increase the training and inference overhead, (ii) overly emphasis on stable model weights corresponding to relations that are invariant across spatial or temporal dimension might cause the loss of informative features.
Generative \cite{generative2023kdd}
CaST \cite{cast2023nips}
}

\textbf{Mixture-of-experts.} The concept of MoE was initially introduced by Jacobs~\etal~\cite{adamoe1991nc} to reduce the negative impact of task interference during training a single neural network. This approach achieved large-scale success when Shazeer~\etal~\cite{sparsemoe2017iclr} refined the gating mechanism with Top-$K$ sparsity constraints. This innovation enabled scaling of the model to billions of parameters with MoE layers integrated, enhancing its capacity remarkably. Subsequent developments in the field have included novel advancements in gating design~\cite{gshard2021iclr,switch2022jmlr}, optimization algorithms~\cite{stmoe2022}, and distributed training frameworks~\cite{switch2022jmlr,deepspeedmoe2022icml}. These advancements have further elevated the capabilities and transferability of MoE-based large language models.
The impressive achievements of MoE in the NLP field have inspired researchers to apply MoE's scalable architecture and conditional computation abilities to fields such as computer vision~\cite{cvmoe2021nips}, multi-modal learning~\cite{multimodalmoe2022nips} and graph learning~\cite{graphmoe2023nips,weakstrongmoe2023}. For instance, GraphMoE~\cite{graphmoe2023nips} employed multiple experts to gather information from node neighbors at different hops, effectively capturing diverse structural knowledge. Mowst~\cite{weakstrongmoe2023} combined simple MLPs and GNNs, leveraging nuanced collaboration between weak and strong experts to improve graph learning. 
In contrast, our approach designed a capable, robust and interpretable spatio-temporal MoE layer with task-customized designs for congestion prediction.

\section{Conclusion}\label{sec:conclusion}
In this study, we propose an effective Congestion Prediction Mixture-of-Experts (\model) to handle the traffic congestion prediction problem. 
We first propose a Mixture of Adaptive Graph Learners (MAGLs) with a tailored sparse gating mechanism and congestion-aware expert biases to effectively capture heterogeneous and evolving traffic patterns. Additionally, we incorporate two specialized experts to capture stable trend and periodic patterns, and adaptively cascade them with MAGLs to boost \model's robustness and interpretability. An ordinal regression strategy is further employed to alleviate over-confidence issues among experts and promote effective collaboration. Extensive experiments on real-world datasets validate \model's superior performance against various baselines. 
Notably, \model has been successfully deployed in DiDi to enhance the reliability of its travel time estimation system. 
In the future, we plan to investigate the utility of \model in a broader range of ride-hailing services, such as route planning, to further enhance operational efficiency and user experiences.


\begin{acks}
This work was supported by the National Natural Science Foundation of China (Grant No.62102110, No.92370204), National Key R\&D Program of China (Grant No.2023YFF0725004), Guangzhou Basic and Applied Basic Research Program (Grant No.2024A04J3279), Education Bureau of Guangzhou Municipality.
\end{acks}

\bibliographystyle{ACM-Reference-Format}
\balance
\bibliography{ref}


\begin{thebibliography}{61}


\ifx \showCODEN    \undefined \def \showCODEN     #1{\unskip}     \fi
\ifx \showDOI      \undefined \def \showDOI       #1{#1}\fi
\ifx \showISBNx    \undefined \def \showISBNx     #1{\unskip}     \fi
\ifx \showISBNxiii \undefined \def \showISBNxiii  #1{\unskip}     \fi
\ifx \showISSN     \undefined \def \showISSN      #1{\unskip}     \fi
\ifx \showLCCN     \undefined \def \showLCCN      #1{\unskip}     \fi
\ifx \shownote     \undefined \def \shownote      #1{#1}          \fi
\ifx \showarticletitle \undefined \def \showarticletitle #1{#1}   \fi
\ifx \showURL      \undefined \def \showURL       {\relax}        \fi
\providecommand\bibfield[2]{#2}
\providecommand\bibinfo[2]{#2}
\providecommand\natexlab[1]{#1}
\providecommand\showeprint[2][]{arXiv:#2}

\bibitem[Akhtar and Moridpour(2021)]%
        {cpsurvey2021jat}
\bibfield{author}{\bibinfo{person}{Mahmuda Akhtar} {and} \bibinfo{person}{Sara Moridpour}.} \bibinfo{year}{2021}\natexlab{}.
\newblock \showarticletitle{A review of traffic congestion prediction using artificial intelligence}.
\newblock \bibinfo{journal}{\emph{Journal of Advanced Transportation}}  \bibinfo{volume}{2021} (\bibinfo{year}{2021}), \bibinfo{pages}{1--18}.
\newblock


\bibitem[Artetxe et~al\mbox{.}(2022)]%
        {nlpmoe2022emnlp}
\bibfield{author}{\bibinfo{person}{Mikel Artetxe}, \bibinfo{person}{Shruti Bhosale}, \bibinfo{person}{Naman Goyal}, \bibinfo{person}{Todor Mihaylov}, \bibinfo{person}{Myle Ott}, \bibinfo{person}{Sam Shleifer}, \bibinfo{person}{Xi~Victoria Lin}, \bibinfo{person}{Jingfei Du}, \bibinfo{person}{Srinivasan Iyer}, \bibinfo{person}{Ramakanth Pasunuru}, \bibinfo{person}{Giridharan Anantharaman}, \bibinfo{person}{Xian Li}, \bibinfo{person}{Shuohui Chen}, \bibinfo{person}{Halil Akin}, \bibinfo{person}{Mandeep Baines}, \bibinfo{person}{Louis Martin}, \bibinfo{person}{Xing Zhou}, \bibinfo{person}{Punit~Singh Koura}, \bibinfo{person}{Brian O'Horo}, \bibinfo{person}{Jeffrey Wang}, \bibinfo{person}{Luke Zettlemoyer}, \bibinfo{person}{Mona~T. Diab}, \bibinfo{person}{Zornitsa Kozareva}, {and} \bibinfo{person}{Veselin Stoyanov}.} \bibinfo{year}{2022}\natexlab{}.
\newblock \showarticletitle{Efficient Large Scale Language Modeling with Mixtures of Experts}. In \bibinfo{booktitle}{\emph{Proceedings of the 2022 Conference on Empirical Methods in Natural Language Processing}}. \bibinfo{pages}{11699--11732}.
\newblock


\bibitem[Bai et~al\mbox{.}(2020)]%
        {agcrn2020nips}
\bibfield{author}{\bibinfo{person}{Lei Bai}, \bibinfo{person}{Lina Yao}, \bibinfo{person}{Can Li}, \bibinfo{person}{Xianzhi Wang}, {and} \bibinfo{person}{Can Wang}.} \bibinfo{year}{2020}\natexlab{}.
\newblock \showarticletitle{Adaptive Graph Convolutional Recurrent Network for Traffic Forecasting}. In \bibinfo{booktitle}{\emph{Advances in Neural Information Processing Systems 33}}.
\newblock


\bibitem[Chen et~al\mbox{.}(2018)]%
        {pcnn2018tits}
\bibfield{author}{\bibinfo{person}{Meng Chen}, \bibinfo{person}{Xiaohui Yu}, {and} \bibinfo{person}{Yang Liu}.} \bibinfo{year}{2018}\natexlab{}.
\newblock \showarticletitle{PCNN: Deep convolutional networks for short-term traffic congestion prediction}.
\newblock \bibinfo{journal}{\emph{IEEE Transactions on Intelligent Transportation Systems}} \bibinfo{volume}{19}, \bibinfo{number}{11} (\bibinfo{year}{2018}), \bibinfo{pages}{3550--3559}.
\newblock


\bibitem[Chen et~al\mbox{.}(2022)]%
        {BAGNN2022icde}
\bibfield{author}{\bibinfo{person}{Zhengyu Chen}, \bibinfo{person}{Teng Xiao}, {and} \bibinfo{person}{Kun Kuang}.} \bibinfo{year}{2022}\natexlab{}.
\newblock \showarticletitle{{BA-GNN:} On Learning Bias-Aware Graph Neural Network}. In \bibinfo{booktitle}{\emph{2022 IEEE 38th International Conference on Data Engineering}}. \bibinfo{pages}{3012--3024}.
\newblock


\bibitem[Cheng et~al\mbox{.}(2018)]%
        {deeptrans2018ijcnn}
\bibfield{author}{\bibinfo{person}{Xingyi Cheng}, \bibinfo{person}{Ruiqing Zhang}, \bibinfo{person}{Jie Zhou}, {and} \bibinfo{person}{Wei Xu}.} \bibinfo{year}{2018}\natexlab{}.
\newblock \showarticletitle{DeepTransport: Learning Spatial-Temporal Dependency for Traffic Condition Forecasting}. In \bibinfo{booktitle}{\emph{2018 International Joint Conference on Neural Networks}}. \bibinfo{pages}{1--8}.
\newblock


\bibitem[Chung et~al\mbox{.}(2014)]%
        {gru2014}
\bibfield{author}{\bibinfo{person}{Junyoung Chung}, \bibinfo{person}{{\c{C}}aglar G{\"{u}}l{\c{c}}ehre}, \bibinfo{person}{KyungHyun Cho}, {and} \bibinfo{person}{Yoshua Bengio}.} \bibinfo{year}{2014}\natexlab{}.
\newblock \showarticletitle{Empirical Evaluation of Gated Recurrent Neural Networks on Sequence Modeling}.
\newblock \bibinfo{journal}{\emph{CoRR}}  \bibinfo{volume}{abs/1412.3555} (\bibinfo{year}{2014}).
\newblock


\bibitem[Cohen et~al\mbox{.}(2009)]%
        {pearson2009}
\bibfield{author}{\bibinfo{person}{Israel Cohen}, \bibinfo{person}{Yiteng Huang}, \bibinfo{person}{Jingdong Chen}, \bibinfo{person}{Jacob Benesty}, \bibinfo{person}{Jacob Benesty}, \bibinfo{person}{Jingdong Chen}, \bibinfo{person}{Yiteng Huang}, {and} \bibinfo{person}{Israel Cohen}.} \bibinfo{year}{2009}\natexlab{}.
\newblock \showarticletitle{Pearson correlation coefficient}.
\newblock \bibinfo{journal}{\emph{Noise Reduction in Speech Processing}} (\bibinfo{year}{2009}), \bibinfo{pages}{1--4}.
\newblock


\bibitem[Deng et~al\mbox{.}(2021)]%
        {stnorm2021kdd}
\bibfield{author}{\bibinfo{person}{Jinliang Deng}, \bibinfo{person}{Xiusi Chen}, \bibinfo{person}{Renhe Jiang}, \bibinfo{person}{Xuan Song}, {and} \bibinfo{person}{Ivor~W. Tsang}.} \bibinfo{year}{2021}\natexlab{}.
\newblock \showarticletitle{ST-Norm: Spatial and Temporal Normalization for Multi-variate Time Series Forecasting}. In \bibinfo{booktitle}{\emph{Proceedings of the 27th ACM SIGKDD Conference on Knowledge Discovery \& Data Mining}}. \bibinfo{pages}{269--278}.
\newblock


\bibitem[Di et~al\mbox{.}(2019)]%
        {cpm2019mdm}
\bibfield{author}{\bibinfo{person}{Xiaolei Di}, \bibinfo{person}{Yu Xiao}, \bibinfo{person}{Chao Zhu}, \bibinfo{person}{Yang Deng}, \bibinfo{person}{Qinpei Zhao}, {and} \bibinfo{person}{Weixiong Rao}.} \bibinfo{year}{2019}\natexlab{}.
\newblock \showarticletitle{Traffic Congestion Prediction by Spatiotemporal Propagation Patterns}. In \bibinfo{booktitle}{\emph{20th IEEE International Conference on Mobile Data Management}}. \bibinfo{pages}{298--303}.
\newblock


\bibitem[Diaz and Marathe(2019)]%
        {softord2019cvpr}
\bibfield{author}{\bibinfo{person}{Raul Diaz} {and} \bibinfo{person}{Amit Marathe}.} \bibinfo{year}{2019}\natexlab{}.
\newblock \showarticletitle{Soft Labels for Ordinal Regression}. In \bibinfo{booktitle}{\emph{2019 {IEEE} Conference on Computer Vision and Pattern Recognition}}. \bibinfo{pages}{4738--4747}.
\newblock


\bibitem[Djahel et~al\mbox{.}(2015)]%
        {congemergency2015}
\bibfield{author}{\bibinfo{person}{Soufiene Djahel}, \bibinfo{person}{Nicolas Smith}, \bibinfo{person}{Shen Wang}, {and} \bibinfo{person}{John Murphy}.} \bibinfo{year}{2015}\natexlab{}.
\newblock \showarticletitle{Reducing emergency services response time in smart cities: An advanced adaptive and fuzzy approach}. In \bibinfo{booktitle}{\emph{2015 IEEE First International Smart Cities Conference}}. \bibinfo{pages}{1--8}.
\newblock


\bibitem[Fang et~al\mbox{.}(2023)]%
        {stwave2023tkde}
\bibfield{author}{\bibinfo{person}{Yuchen Fang}, \bibinfo{person}{Yanjun Qin}, \bibinfo{person}{Haiyong Luo}, \bibinfo{person}{Fang Zhao}, {and} \bibinfo{person}{Kai Zheng}.} \bibinfo{year}{2023}\natexlab{}.
\newblock \showarticletitle{STWave+: A Multi-Scale Efficient Spectral Graph Attention Network With Long-Term Trends for Disentangled Traffic Flow Forecasting}.
\newblock \bibinfo{journal}{\emph{IEEE Transactions on Knowledge and Data Engineering}} (\bibinfo{year}{2023}).
\newblock


\bibitem[Fedus et~al\mbox{.}(2022a)]%
        {moesurvey2022}
\bibfield{author}{\bibinfo{person}{William Fedus}, \bibinfo{person}{Jeff Dean}, {and} \bibinfo{person}{Barret Zoph}.} \bibinfo{year}{2022}\natexlab{a}.
\newblock \showarticletitle{A Review of Sparse Expert Models in Deep Learning}.
\newblock \bibinfo{journal}{\emph{CoRR}}  \bibinfo{volume}{abs/2209.01667} (\bibinfo{year}{2022}).
\newblock


\bibitem[Fedus et~al\mbox{.}(2022b)]%
        {switch2022jmlr}
\bibfield{author}{\bibinfo{person}{William Fedus}, \bibinfo{person}{Barret Zoph}, {and} \bibinfo{person}{Noam Shazeer}.} \bibinfo{year}{2022}\natexlab{b}.
\newblock \showarticletitle{Switch Transformers: Scaling to Trillion Parameter Models with Simple and Efficient Sparsity}.
\newblock \bibinfo{journal}{\emph{Journal of Machine Learning Research}}  \bibinfo{volume}{23} (\bibinfo{year}{2022}), \bibinfo{pages}{120:1--120:39}.
\newblock


\bibitem[Glorot et~al\mbox{.}(2011)]%
        {softplus2011aistat}
\bibfield{author}{\bibinfo{person}{Xavier Glorot}, \bibinfo{person}{Antoine Bordes}, {and} \bibinfo{person}{Yoshua Bengio}.} \bibinfo{year}{2011}\natexlab{}.
\newblock \showarticletitle{Deep Sparse Rectifier Neural Networks}. In \bibinfo{booktitle}{\emph{Proceedings of the 14th International Conference on Artificial Intelligence and Statistics}}, Vol.~\bibinfo{volume}{15}. \bibinfo{pages}{315--323}.
\newblock


\bibitem[Gong et~al\mbox{.}(2022)]%
        {ranksim2022icml}
\bibfield{author}{\bibinfo{person}{Yu Gong}, \bibinfo{person}{Greg Mori}, {and} \bibinfo{person}{Frederick Tung}.} \bibinfo{year}{2022}\natexlab{}.
\newblock \showarticletitle{RankSim: Ranking Similarity Regularization for Deep Imbalanced Regression}. In \bibinfo{booktitle}{\emph{Proceedings of the 39th International Conference on Machine Learning}}, Vol.~\bibinfo{volume}{162}. \bibinfo{pages}{7634--7649}.
\newblock


\bibitem[Guo et~al\mbox{.}(2022)]%
        {astgnn2022tkde}
\bibfield{author}{\bibinfo{person}{Shengnan Guo}, \bibinfo{person}{Youfang Lin}, \bibinfo{person}{Huaiyu Wan}, \bibinfo{person}{Xiucheng Li}, {and} \bibinfo{person}{Gao Cong}.} \bibinfo{year}{2022}\natexlab{}.
\newblock \showarticletitle{Learning Dynamics and Heterogeneity of Spatial-Temporal Graph Data for Traffic Forecasting}.
\newblock \bibinfo{journal}{\emph{IEEE Transactions on Knowledge and Data Engineering}} \bibinfo{volume}{34}, \bibinfo{number}{11} (\bibinfo{year}{2022}), \bibinfo{pages}{5415--5428}.
\newblock


\bibitem[Han et~al\mbox{.}(2023)]%
        {ieta2023kdd}
\bibfield{author}{\bibinfo{person}{Jindong Han}, \bibinfo{person}{Hao Liu}, \bibinfo{person}{Shui Liu}, \bibinfo{person}{Xi Chen}, \bibinfo{person}{Naiqiang Tan}, \bibinfo{person}{Hua Chai}, {and} \bibinfo{person}{Hui Xiong}.} \bibinfo{year}{2023}\natexlab{}.
\newblock \showarticletitle{iETA: {A} Robust and Scalable Incremental Learning Framework for Time-of-Arrival Estimation}. In \bibinfo{booktitle}{\emph{Proceedings of the 29th ACM SIGKDD Conference on Knowledge Discovery \& Data Mining}}. \bibinfo{pages}{4100--4111}.
\newblock


\bibitem[Han et~al\mbox{.}(2024)]%
        {bigst2024vldb}
\bibfield{author}{\bibinfo{person}{Jindong Han}, \bibinfo{person}{Weijia Zhang}, \bibinfo{person}{Hao Liu}, \bibinfo{person}{Tao Tao}, \bibinfo{person}{Naiqiang Tan}, {and} \bibinfo{person}{Hui Xiong}.} \bibinfo{year}{2024}\natexlab{}.
\newblock \showarticletitle{BigST: Linear Complexity Spatio-Temporal Graph Neural Network for Traffic Forecasting on Large-Scale Road Networks}.
\newblock \bibinfo{journal}{\emph{Proceedings of the VLDB Endowment}} \bibinfo{volume}{17}, \bibinfo{number}{5} (\bibinfo{year}{2024}), \bibinfo{pages}{1081--1090}.
\newblock


\bibitem[Heil and Walnut(1989)]%
        {wavelet1989siam}
\bibfield{author}{\bibinfo{person}{Christopher~E Heil} {and} \bibinfo{person}{David~F Walnut}.} \bibinfo{year}{1989}\natexlab{}.
\newblock \showarticletitle{Continuous and discrete wavelet transforms}.
\newblock \bibinfo{journal}{\emph{SIAM Rev.}} \bibinfo{volume}{31}, \bibinfo{number}{4} (\bibinfo{year}{1989}), \bibinfo{pages}{628--666}.
\newblock


\bibitem[Inoue et~al\mbox{.}(2016)]%
        {freqpat_cong2016itsc}
\bibfield{author}{\bibinfo{person}{Ryo Inoue}, \bibinfo{person}{Akihisa Miyashita}, {and} \bibinfo{person}{Masatoshi Sugita}.} \bibinfo{year}{2016}\natexlab{}.
\newblock \showarticletitle{Mining spatio-temporal patterns of congested traffic in urban areas from traffic sensor data}. In \bibinfo{booktitle}{\emph{19th IEEE Intelligent Transportation Systems Conference}}. \bibinfo{pages}{731--736}.
\newblock


\bibitem[Jacobs et~al\mbox{.}(1991)]%
        {adamoe1991nc}
\bibfield{author}{\bibinfo{person}{Robert~A. Jacobs}, \bibinfo{person}{Michael~I. Jordan}, \bibinfo{person}{Steven~J. Nowlan}, {and} \bibinfo{person}{Geoffrey~E. Hinton}.} \bibinfo{year}{1991}\natexlab{}.
\newblock \showarticletitle{Adaptive Mixtures of Local Experts}.
\newblock \bibinfo{journal}{\emph{Neural Computing}} \bibinfo{volume}{3}, \bibinfo{number}{1} (\bibinfo{year}{1991}), \bibinfo{pages}{79--87}.
\newblock


\bibitem[Jin et~al\mbox{.}(2023)]%
        {congevent2023aaai}
\bibfield{author}{\bibinfo{person}{Guangyin Jin}, \bibinfo{person}{Lingbo Liu}, \bibinfo{person}{Fuxian Li}, {and} \bibinfo{person}{Jincai Huang}.} \bibinfo{year}{2023}\natexlab{}.
\newblock \showarticletitle{Spatio-Temporal Graph Neural Point Process for Traffic Congestion Event Prediction}. In \bibinfo{booktitle}{\emph{Proceedings of the 37th AAAI Conference on Artificial Intelligence}}. \bibinfo{pages}{14268--14276}.
\newblock


\bibitem[Kim and Wang(2016)]%
        {bayescong2016trr}
\bibfield{author}{\bibinfo{person}{Jiwon Kim} {and} \bibinfo{person}{Guangxing Wang}.} \bibinfo{year}{2016}\natexlab{}.
\newblock \showarticletitle{Diagnosis and prediction of traffic congestion on urban road networks using Bayesian networks}.
\newblock \bibinfo{journal}{\emph{Transportation Research Record}} \bibinfo{volume}{2595}, \bibinfo{number}{1} (\bibinfo{year}{2016}), \bibinfo{pages}{108--118}.
\newblock


\bibitem[Kumar and Raubal(2021)]%
        {cpsurvey2021trpc}
\bibfield{author}{\bibinfo{person}{Nishant Kumar} {and} \bibinfo{person}{Martin Raubal}.} \bibinfo{year}{2021}\natexlab{}.
\newblock \showarticletitle{Applications of deep learning in congestion detection, prediction and alleviation: A survey}.
\newblock \bibinfo{journal}{\emph{Transportation Research Part C: Emerging Technologies}}  \bibinfo{volume}{133} (\bibinfo{year}{2021}), \bibinfo{pages}{103432}.
\newblock


\bibitem[Lepikhin et~al\mbox{.}(2021)]%
        {gshard2021iclr}
\bibfield{author}{\bibinfo{person}{Dmitry Lepikhin}, \bibinfo{person}{HyoukJoong Lee}, \bibinfo{person}{Yuanzhong Xu}, \bibinfo{person}{Dehao Chen}, \bibinfo{person}{Orhan Firat}, \bibinfo{person}{Yanping Huang}, \bibinfo{person}{Maxim Krikun}, \bibinfo{person}{Noam Shazeer}, {and} \bibinfo{person}{Zhifeng Chen}.} \bibinfo{year}{2021}\natexlab{}.
\newblock \showarticletitle{GShard: Scaling Giant Models with Conditional Computation and Automatic Sharding}. In \bibinfo{booktitle}{\emph{9th International Conference on Learning Representations}}.
\newblock


\bibitem[Li et~al\mbox{.}(2016)]%
        {clustercong2016}
\bibfield{author}{\bibinfo{person}{Fuliang Li}, \bibinfo{person}{Junfeng Gong}, \bibinfo{person}{Yunyi Liang}, {and} \bibinfo{person}{Jiali Zhou}.} \bibinfo{year}{2016}\natexlab{}.
\newblock \showarticletitle{Real-time congestion prediction for urban arterials using adaptive data-driven methods}.
\newblock \bibinfo{journal}{\emph{Multimedia Tools and Applications}}  \bibinfo{volume}{75} (\bibinfo{year}{2016}), \bibinfo{pages}{17573--17592}.
\newblock


\bibitem[Li et~al\mbox{.}(2023b)]%
        {psestgcn2023sigspatial}
\bibfield{author}{\bibinfo{person}{Fuxian Li}, \bibinfo{person}{Huan Yan}, \bibinfo{person}{Hongjie Sui}, \bibinfo{person}{Deng Wang}, \bibinfo{person}{Fan Zuo}, \bibinfo{person}{Yue Liu}, \bibinfo{person}{Yong Li}, {and} \bibinfo{person}{Depeng Jin}.} \bibinfo{year}{2023}\natexlab{b}.
\newblock \showarticletitle{Periodic Shift and Event-aware Spatio-Temporal Graph Convolutional Network for Traffic Congestion Prediction}. In \bibinfo{booktitle}{\emph{Proceedings of the 31st {ACM} International Conference on Advances in Geographic Information Systems}}. \bibinfo{pages}{50:1--50:10}.
\newblock


\bibitem[Li et~al\mbox{.}(2023a)]%
        {stmoe2023cikm}
\bibfield{author}{\bibinfo{person}{Shuhao Li}, \bibinfo{person}{Yue Cui}, \bibinfo{person}{Yan Zhao}, \bibinfo{person}{Weidong Yang}, \bibinfo{person}{Ruiyuan Zhang}, {and} \bibinfo{person}{Xiaofang Zhou}.} \bibinfo{year}{2023}\natexlab{a}.
\newblock \showarticletitle{ST-MoE: Spatio-Temporal Mixture-of-Experts for Debiasing in Traffic Prediction}. In \bibinfo{booktitle}{\emph{Proceedings of the 32nd {ACM} International Conference on Information {\&} Knowledge Management}}. \bibinfo{pages}{1208--1217}.
\newblock


\bibitem[Li et~al\mbox{.}(2018)]%
        {dcrnn2018iclr}
\bibfield{author}{\bibinfo{person}{Yaguang Li}, \bibinfo{person}{Rose Yu}, \bibinfo{person}{Cyrus Shahabi}, {and} \bibinfo{person}{Yan Liu}.} \bibinfo{year}{2018}\natexlab{}.
\newblock \showarticletitle{Diffusion Convolutional Recurrent Neural Network: Data-Driven Traffic Forecasting}. In \bibinfo{booktitle}{\emph{6th International Conference on Learning Representations}}.
\newblock


\bibitem[Liu et~al\mbox{.}(2022)]%
        {ASTFA2022nips}
\bibfield{author}{\bibinfo{person}{Fan Liu}, \bibinfo{person}{Hao Liu}, {and} \bibinfo{person}{Wenzhao Jiang}.} \bibinfo{year}{2022}\natexlab{}.
\newblock \showarticletitle{Practical Adversarial Attacks on Spatiotemporal Traffic Forecasting Models}. In \bibinfo{booktitle}{\emph{Advances in Neural Information Processing Systems 35}}. \bibinfo{pages}{19035--19047}.
\newblock


\bibitem[Liu et~al\mbox{.}(2023a)]%
        {staeformer2023cikm}
\bibfield{author}{\bibinfo{person}{Hangchen Liu}, \bibinfo{person}{Zheng Dong}, \bibinfo{person}{Renhe Jiang}, \bibinfo{person}{Jiewen Deng}, \bibinfo{person}{Jinliang Deng}, \bibinfo{person}{Quanjun Chen}, {and} \bibinfo{person}{Xuan Song}.} \bibinfo{year}{2023}\natexlab{a}.
\newblock \showarticletitle{Spatio-temporal adaptive embedding makes vanilla transformer sota for traffic forecasting}. In \bibinfo{booktitle}{\emph{Proceedings of the 32nd {ACM} International Conference on Information {\&} Knowledge Management}}. \bibinfo{pages}{4125--4129}.
\newblock


\bibitem[Liu et~al\mbox{.}(2023b)]%
        {probtte2023kdd}
\bibfield{author}{\bibinfo{person}{Hao Liu}, \bibinfo{person}{Wenzhao Jiang}, \bibinfo{person}{Shui Liu}, {and} \bibinfo{person}{Xi Chen}.} \bibinfo{year}{2023}\natexlab{b}.
\newblock \showarticletitle{Uncertainty-Aware Probabilistic Travel Time Prediction for On-Demand Ride-Hailing at DiDi}. In \bibinfo{booktitle}{\emph{Proceedings of the 29th ACM SIGKDD Conference on Knowledge Discovery \& Data Mining}}. \bibinfo{pages}{4516--4526}.
\newblock


\bibitem[Liu et~al\mbox{.}(2020)]%
        {polestar2020kdd}
\bibfield{author}{\bibinfo{person}{Hao Liu}, \bibinfo{person}{Ying Li}, \bibinfo{person}{Yanjie Fu}, \bibinfo{person}{Huaibo Mei}, \bibinfo{person}{Jingbo Zhou}, \bibinfo{person}{Xu Ma}, {and} \bibinfo{person}{Hui Xiong}.} \bibinfo{year}{2020}\natexlab{}.
\newblock \showarticletitle{Polestar: An Intelligent, Efficient and National-Wide Public Transportation Routing Engine}. In \bibinfo{booktitle}{\emph{Proceedings of the 26th ACM SIGKDD International Conference on Knowledge Discovery \& Data Mining}}. \bibinfo{pages}{2321--2329}.
\newblock


\bibitem[Lv et~al\mbox{.}(2021)]%
        {revisit_hgnn2021kdd}
\bibfield{author}{\bibinfo{person}{Qingsong Lv}, \bibinfo{person}{Ming Ding}, \bibinfo{person}{Qiang Liu}, \bibinfo{person}{Yuxiang Chen}, \bibinfo{person}{Wenzheng Feng}, \bibinfo{person}{Siming He}, \bibinfo{person}{Chang Zhou}, \bibinfo{person}{Jianguo Jiang}, \bibinfo{person}{Yuxiao Dong}, {and} \bibinfo{person}{Jie Tang}.} \bibinfo{year}{2021}\natexlab{}.
\newblock \showarticletitle{Are we really making much progress?: Revisiting, benchmarking and refining heterogeneous graph neural networks}. In \bibinfo{booktitle}{\emph{Proceedings of the 27th ACM SIGKDD Conference on Knowledge Discovery \& Data Mining}}. \bibinfo{pages}{1150--1160}.
\newblock


\bibitem[Ma et~al\mbox{.}(2015)]%
        {rbmrnn2015plos}
\bibfield{author}{\bibinfo{person}{Xiaolei Ma}, \bibinfo{person}{Haiyang Yu}, \bibinfo{person}{Yunpeng Wang}, {and} \bibinfo{person}{Yinhai Wang}.} \bibinfo{year}{2015}\natexlab{}.
\newblock \showarticletitle{Large-scale transportation network congestion evolution prediction using deep learning theory}.
\newblock \bibinfo{journal}{\emph{PloS One}} \bibinfo{volume}{10}, \bibinfo{number}{3} (\bibinfo{year}{2015}), \bibinfo{pages}{e0119044}.
\newblock


\bibitem[Mustafa et~al\mbox{.}(2022)]%
        {multimodalmoe2022nips}
\bibfield{author}{\bibinfo{person}{Basil Mustafa}, \bibinfo{person}{Carlos Riquelme}, \bibinfo{person}{Joan Puigcerver}, \bibinfo{person}{Rodolphe Jenatton}, {and} \bibinfo{person}{Neil Houlsby}.} \bibinfo{year}{2022}\natexlab{}.
\newblock \showarticletitle{Multimodal Contrastive Learning with LIMoE: the Language-Image Mixture of Experts}. In \bibinfo{booktitle}{\emph{Advances in Neural Information Processing Systems 35}}.
\newblock


\bibitem[Park and Kim(2022)]%
        {vitwork2022iclr}
\bibfield{author}{\bibinfo{person}{Namuk Park} {and} \bibinfo{person}{Songkuk Kim}.} \bibinfo{year}{2022}\natexlab{}.
\newblock \showarticletitle{How Do Vision Transformers Work?}. In \bibinfo{booktitle}{\emph{10th International Conference on Learning Representations}}.
\newblock


\bibitem[Qi and Ishak(2014)]%
        {hmmcong2014trpc}
\bibfield{author}{\bibinfo{person}{Yan Qi} {and} \bibinfo{person}{Sherif Ishak}.} \bibinfo{year}{2014}\natexlab{}.
\newblock \showarticletitle{A Hidden Markov Model for short term prediction of traffic conditions on freeways}.
\newblock \bibinfo{journal}{\emph{Transportation Research Part C: Emerging Technologies}}  \bibinfo{volume}{43} (\bibinfo{year}{2014}), \bibinfo{pages}{95--111}.
\newblock


\bibitem[Rajbhandari et~al\mbox{.}(2022)]%
        {deepspeedmoe2022icml}
\bibfield{author}{\bibinfo{person}{Samyam Rajbhandari}, \bibinfo{person}{Conglong Li}, \bibinfo{person}{Zhewei Yao}, \bibinfo{person}{Minjia Zhang}, \bibinfo{person}{Reza~Yazdani Aminabadi}, \bibinfo{person}{Ammar~Ahmad Awan}, \bibinfo{person}{Jeff Rasley}, {and} \bibinfo{person}{Yuxiong He}.} \bibinfo{year}{2022}\natexlab{}.
\newblock \showarticletitle{DeepSpeed-MoE: Advancing Mixture-of-Experts Inference and Training to Power Next-Generation {AI} Scale}. In \bibinfo{booktitle}{\emph{Proceedings of the 39th International Conference on Machine Learning}}, Vol.~\bibinfo{volume}{162}. \bibinfo{pages}{18332--18346}.
\newblock


\bibitem[Riquelme et~al\mbox{.}(2021)]%
        {cvmoe2021nips}
\bibfield{author}{\bibinfo{person}{Carlos Riquelme}, \bibinfo{person}{Joan Puigcerver}, \bibinfo{person}{Basil Mustafa}, \bibinfo{person}{Maxim Neumann}, \bibinfo{person}{Rodolphe Jenatton}, \bibinfo{person}{Andr{\'{e}}~Susano Pinto}, \bibinfo{person}{Daniel Keysers}, {and} \bibinfo{person}{Neil Houlsby}.} \bibinfo{year}{2021}\natexlab{}.
\newblock \showarticletitle{Scaling Vision with Sparse Mixture of Experts}. In \bibinfo{booktitle}{\emph{Advances in Neural Information Processing Systems 34}}. \bibinfo{pages}{8583--8595}.
\newblock


\bibitem[Rudin(2019)]%
        {inherentinter2019nature}
\bibfield{author}{\bibinfo{person}{Cynthia Rudin}.} \bibinfo{year}{2019}\natexlab{}.
\newblock \showarticletitle{Stop explaining black box machine learning models for high stakes decisions and use interpretable models instead}.
\newblock \bibinfo{journal}{\emph{Nature Machine Intelligence}} \bibinfo{volume}{1}, \bibinfo{number}{5} (\bibinfo{year}{2019}), \bibinfo{pages}{206--215}.
\newblock


\bibitem[Shao et~al\mbox{.}(2022)]%
        {stid2022cikm}
\bibfield{author}{\bibinfo{person}{Zezhi Shao}, \bibinfo{person}{Zhao Zhang}, \bibinfo{person}{Fei Wang}, \bibinfo{person}{Wei Wei}, {and} \bibinfo{person}{Yongjun Xu}.} \bibinfo{year}{2022}\natexlab{}.
\newblock \showarticletitle{Spatial-Temporal Identity: {A} Simple yet Effective Baseline for Multivariate Time Series Forecasting}. In \bibinfo{booktitle}{\emph{Proceedings of the 31st {ACM} International Conference on Information {\&} Knowledge Management}}. \bibinfo{pages}{4454--4458}.
\newblock


\bibitem[Shazeer et~al\mbox{.}(2017)]%
        {sparsemoe2017iclr}
\bibfield{author}{\bibinfo{person}{Noam Shazeer}, \bibinfo{person}{Azalia Mirhoseini}, \bibinfo{person}{Krzysztof Maziarz}, \bibinfo{person}{Andy Davis}, \bibinfo{person}{Quoc~V. Le}, \bibinfo{person}{Geoffrey~E. Hinton}, {and} \bibinfo{person}{Jeff Dean}.} \bibinfo{year}{2017}\natexlab{}.
\newblock \showarticletitle{Outrageously Large Neural Networks: The Sparsely-Gated Mixture-of-Experts Layer}. In \bibinfo{booktitle}{\emph{5th International Conference on Learning Representations}}.
\newblock


\bibitem[Sun et~al\mbox{.}(2021)]%
        {dmlm2021ijcai}
\bibfield{author}{\bibinfo{person}{Yidan Sun}, \bibinfo{person}{Guiyuan Jiang}, \bibinfo{person}{Siew~Kei Lam}, {and} \bibinfo{person}{Peilan He}.} \bibinfo{year}{2021}\natexlab{}.
\newblock \showarticletitle{Predicting Traffic Congestion Evolution: {A} Deep Meta Learning Approach}. In \bibinfo{booktitle}{\emph{Proceedings of the 30th International Joint Conference on Artificial Intelligence}}. \bibinfo{pages}{3031--3037}.
\newblock


\bibitem[Vaswani et~al\mbox{.}(2017)]%
        {transformer2017nips}
\bibfield{author}{\bibinfo{person}{Ashish Vaswani}, \bibinfo{person}{Noam Shazeer}, \bibinfo{person}{Niki Parmar}, \bibinfo{person}{Jakob Uszkoreit}, \bibinfo{person}{Llion Jones}, \bibinfo{person}{Aidan~N. Gomez}, \bibinfo{person}{Lukasz Kaiser}, {and} \bibinfo{person}{Illia Polosukhin}.} \bibinfo{year}{2017}\natexlab{}.
\newblock \showarticletitle{Attention is All you Need}. In \bibinfo{booktitle}{\emph{Advances in Neural Information Processing Systems 30}}. \bibinfo{pages}{5998--6008}.
\newblock


\bibitem[Velickovic et~al\mbox{.}(2018)]%
        {gat2018iclr}
\bibfield{author}{\bibinfo{person}{Petar Velickovic}, \bibinfo{person}{Guillem Cucurull}, \bibinfo{person}{Arantxa Casanova}, \bibinfo{person}{Adriana Romero}, \bibinfo{person}{Pietro Li{\`{o}}}, {and} \bibinfo{person}{Yoshua Bengio}.} \bibinfo{year}{2018}\natexlab{}.
\newblock \showarticletitle{Graph Attention Networks}. In \bibinfo{booktitle}{\emph{6th International Conference on Learning Representations}}.
\newblock


\bibitem[Wang et~al\mbox{.}(2023a)]%
        {graphmoe2023nips}
\bibfield{author}{\bibinfo{person}{Haotao Wang}, \bibinfo{person}{Ziyu Jiang}, \bibinfo{person}{Yuning You}, \bibinfo{person}{Yan Han}, \bibinfo{person}{Gaowen Liu}, \bibinfo{person}{Jayanth Srinivasa}, \bibinfo{person}{Ramana Kompella}, {and} \bibinfo{person}{Zhangyang Wang}.} \bibinfo{year}{2023}\natexlab{a}.
\newblock \showarticletitle{Graph Mixture of Experts: Learning on Large-Scale Graphs with Explicit Diversity Modeling}. In \bibinfo{booktitle}{\emph{Advances in Neural Information Processing Systems 36}}.
\newblock


\bibitem[Wang et~al\mbox{.}(2023b)]%
        {sttf2023ijcis}
\bibfield{author}{\bibinfo{person}{Xing Wang}, \bibinfo{person}{Ruihao Zeng}, \bibinfo{person}{Fumin Zou}, \bibinfo{person}{Lyuchao Liao}, {and} \bibinfo{person}{Faliang Huang}.} \bibinfo{year}{2023}\natexlab{b}.
\newblock \showarticletitle{{STTF:} An Efficient Transformer Model for Traffic Congestion Prediction}.
\newblock \bibinfo{journal}{\emph{International Journal of Computational Intelligence Systems}} \bibinfo{volume}{16}, \bibinfo{number}{1} (\bibinfo{year}{2023}), \bibinfo{pages}{2}.
\newblock


\bibitem[Wei et~al\mbox{.}(2022)]%
        {congtransit2022}
\bibfield{author}{\bibinfo{person}{Keji Wei}, \bibinfo{person}{Vikrant Vaze}, {and} \bibinfo{person}{Alexandre Jacquillat}.} \bibinfo{year}{2022}\natexlab{}.
\newblock \showarticletitle{Transit planning optimization under ride-hailing competition and traffic congestion}.
\newblock \bibinfo{journal}{\emph{Transportation Science}} \bibinfo{volume}{56}, \bibinfo{number}{3} (\bibinfo{year}{2022}), \bibinfo{pages}{725--749}.
\newblock


\bibitem[Wu et~al\mbox{.}(2020)]%
        {mtgnn2020kdd}
\bibfield{author}{\bibinfo{person}{Zonghan Wu}, \bibinfo{person}{Shirui Pan}, \bibinfo{person}{Guodong Long}, \bibinfo{person}{Jing Jiang}, \bibinfo{person}{Xiaojun Chang}, {and} \bibinfo{person}{Chengqi Zhang}.} \bibinfo{year}{2020}\natexlab{}.
\newblock \showarticletitle{Connecting the Dots: Multivariate Time Series Forecasting with Graph Neural Networks}. In \bibinfo{booktitle}{\emph{Proceedings of the 26th ACM SIGKDD International Conference on Knowledge Discovery \& Data Mining}}. \bibinfo{pages}{753--763}.
\newblock


\bibitem[Wu et~al\mbox{.}(2019)]%
        {gwnet2019ijcai}
\bibfield{author}{\bibinfo{person}{Zonghan Wu}, \bibinfo{person}{Shirui Pan}, \bibinfo{person}{Guodong Long}, \bibinfo{person}{Jing Jiang}, {and} \bibinfo{person}{Chengqi Zhang}.} \bibinfo{year}{2019}\natexlab{}.
\newblock \showarticletitle{Graph WaveNet for Deep Spatial-Temporal Graph Modeling}. In \bibinfo{booktitle}{\emph{Proceedings of the 28th International Joint Conference on Artificial Intelligence}}. \bibinfo{pages}{1907--1913}.
\newblock


\bibitem[Xia et~al\mbox{.}(2022)]%
        {dutraffic2022cikm}
\bibfield{author}{\bibinfo{person}{Deguo Xia}, \bibinfo{person}{Xiyan Liu}, \bibinfo{person}{Wei Zhang}, \bibinfo{person}{Hui Zhao}, \bibinfo{person}{Chengzhou Li}, \bibinfo{person}{Weiming Zhang}, \bibinfo{person}{Jizhou Huang}, {and} \bibinfo{person}{Haifeng Wang}.} \bibinfo{year}{2022}\natexlab{}.
\newblock \showarticletitle{DuTraffic: Live Traffic Condition Prediction with Trajectory Data and Street Views at Baidu Maps}. In \bibinfo{booktitle}{\emph{Proceedings of the 31st {ACM} International Conference on Information {\&} Knowledge Management}}. \bibinfo{pages}{3575--3583}.
\newblock


\bibitem[Yang et~al\mbox{.}(2021)]%
        {imbreg2021icml}
\bibfield{author}{\bibinfo{person}{Yuzhe Yang}, \bibinfo{person}{Kaiwen Zha}, \bibinfo{person}{Ying{-}Cong Chen}, \bibinfo{person}{Hao Wang}, {and} \bibinfo{person}{Dina Katabi}.} \bibinfo{year}{2021}\natexlab{}.
\newblock \showarticletitle{Delving into Deep Imbalanced Regression}. In \bibinfo{booktitle}{\emph{Proceedings of the 38th International Conference on Machine Learning}}, Vol.~\bibinfo{volume}{139}. \bibinfo{pages}{11842--11851}.
\newblock


\bibitem[Zaharia et~al\mbox{.}(2016)]%
        {zaharia2016apache}
\bibfield{author}{\bibinfo{person}{Matei Zaharia}, \bibinfo{person}{Reynold~S Xin}, \bibinfo{person}{Patrick Wendell}, \bibinfo{person}{Tathagata Das}, \bibinfo{person}{Michael Armbrust}, \bibinfo{person}{Ankur Dave}, \bibinfo{person}{Xiangrui Meng}, \bibinfo{person}{Josh Rosen}, \bibinfo{person}{Shivaram Venkataraman}, \bibinfo{person}{Michael~J Franklin}, {et~al\mbox{.}}} \bibinfo{year}{2016}\natexlab{}.
\newblock \showarticletitle{Apache spark: a unified engine for big data processing}.
\newblock \bibinfo{journal}{\emph{Commun. ACM}} \bibinfo{volume}{59}, \bibinfo{number}{11} (\bibinfo{year}{2016}), \bibinfo{pages}{56--65}.
\newblock


\bibitem[Zeng et~al\mbox{.}(2023)]%
        {weakstrongmoe2023}
\bibfield{author}{\bibinfo{person}{Hanqing Zeng}, \bibinfo{person}{Hanjia Lyu}, \bibinfo{person}{Diyi Hu}, \bibinfo{person}{Yinglong Xia}, {and} \bibinfo{person}{Jiebo Luo}.} \bibinfo{year}{2023}\natexlab{}.
\newblock \showarticletitle{Mixture of Weak {\&} Strong Experts on Graphs}.
\newblock \bibinfo{journal}{\emph{CoRR}}  \bibinfo{volume}{abs/2311.05185} (\bibinfo{year}{2023}).
\newblock


\bibitem[Zhang et~al\mbox{.}(2021)]%
        {mugrep2021kdd}
\bibfield{author}{\bibinfo{person}{Weijia Zhang}, \bibinfo{person}{Hao Liu}, \bibinfo{person}{Lijun Zha}, \bibinfo{person}{Hengshu Zhu}, \bibinfo{person}{Ji Liu}, \bibinfo{person}{Dejing Dou}, {and} \bibinfo{person}{Hui Xiong}.} \bibinfo{year}{2021}\natexlab{}.
\newblock \showarticletitle{MugRep: {A} Multi-Task Hierarchical Graph Representation Learning Framework for Real Estate Appraisal}. In \bibinfo{booktitle}{\emph{Proceedings of the 27th ACM SIGKDD Conference on Knowledge Discovery \& Data Mining}}. \bibinfo{pages}{3937--3947}.
\newblock


\bibitem[Zhang et~al\mbox{.}(2023)]%
        {robustmoecv2023iccv}
\bibfield{author}{\bibinfo{person}{Yihua Zhang}, \bibinfo{person}{Ruisi Cai}, \bibinfo{person}{Tianlong Chen}, \bibinfo{person}{Guanhua Zhang}, \bibinfo{person}{Huan Zhang}, \bibinfo{person}{Pin{-}Yu Chen}, \bibinfo{person}{Shiyu Chang}, \bibinfo{person}{Zhangyang Wang}, {and} \bibinfo{person}{Sijia Liu}.} \bibinfo{year}{2023}\natexlab{}.
\newblock \showarticletitle{Robust Mixture-of-Expert Training for Convolutional Neural Networks}. In \bibinfo{booktitle}{\emph{2023 {IEEE/CVF} International Conference on Computer Vision}}. \bibinfo{pages}{90--101}.
\newblock


\bibitem[Zheng et~al\mbox{.}(2014)]%
        {urbancomputing2014tist}
\bibfield{author}{\bibinfo{person}{Yu Zheng}, \bibinfo{person}{Licia Capra}, \bibinfo{person}{Ouri Wolfson}, {and} \bibinfo{person}{Hai Yang}.} \bibinfo{year}{2014}\natexlab{}.
\newblock \showarticletitle{Urban Computing: Concepts, Methodologies, and Applications}.
\newblock \bibinfo{journal}{\emph{ACM Transactions on Intelligent Systems and Technology}} \bibinfo{volume}{5}, \bibinfo{number}{3} (\bibinfo{year}{2014}), \bibinfo{pages}{38:1--38:55}.
\newblock


\bibitem[Zoph et~al\mbox{.}(2022)]%
        {stmoe2022}
\bibfield{author}{\bibinfo{person}{Barret Zoph}, \bibinfo{person}{Irwan Bello}, \bibinfo{person}{Sameer Kumar}, \bibinfo{person}{Nan Du}, \bibinfo{person}{Yanping Huang}, \bibinfo{person}{Jeff Dean}, \bibinfo{person}{Noam Shazeer}, {and} \bibinfo{person}{William Fedus}.} \bibinfo{year}{2022}\natexlab{}.
\newblock \showarticletitle{Designing Effective Sparse Expert Models}.
\newblock \bibinfo{journal}{\emph{CoRR}}  \bibinfo{volume}{abs/2202.08906} (\bibinfo{year}{2022}).
\newblock


\end{thebibliography}

\appendix


\section{Model Details}
\subsection{Gated Temporal Convolutional Networks} \label{app:tcn}
The gated TCN 
comprises of dilated causal convolution layers and an output gate.
The causal convolution, an extension of 1D convolution, preserves temporal ordering by only convolving over the preceding time intervals. 
The dilated causal convolution stacks multiple causal convolutions with exponentially increasing sliding steps, enabling efficient extraction of multi-level temporal patterns over a much wider receptive field.
Formally, a dilated causal convolution operation for a 1D sequence $\mathbf{x}$ at time $t$ is defined as 
\begin{align}
    (\mathbf{x} *_d \mathbf{\Theta})(t) = \sum_{s=0}^{S-1} \mathbf{\Theta}(s) \cdot \mathbf{x}(t - d\cdot s),
\end{align}
where $*_d$ denotes the dilated convolution, $\Theta(\cdot)$ is the learnable convolution filter, $d$ is the dilation factor, and $S$ is the size of filter.
Furthermore, a output gate is incorporated to control the ratios of information flowing through dilated convolution layers. 
Overall, the short-term temporal encodings of link $v_i$ at time interval $t$ are computed as
\begin{align}
    {\mathbf{H}_i^t}' = g(\mathbf{H}_i^t *_d \mathbf{\Theta}_1 + \mathbf{b}) \odot \sigma(\mathbf{H}_i^t *_d \mathbf{\Theta}_2 + \mathbf{c}),
\end{align}
where $\mathbf{b}$ and $\mathbf{c}$ are learnable bias terms, $g(\cdot)$ is an activation function and $\sigma(\cdot)$ is the sigmoid function.

\subsection{Discrete Wavelet Transform} \label{app:dwt}
The Discrete Wavelet Transform~(DWT) is a powerful mathematical tool that allows us to analyze various frequency components of a signal with a resolution matched to each scale.
It decomposes the original signal $x(t)$ into approximation coefficients $A$ and detail coefficients $\{D_i\}$ by convolving $x(t)$ with low-pass ($f_{l}$) and high-pass ($f_{h}$) filters respectively, followed by downsampling,
\begin{align}
    A(k) = (x * f_l)(2k), \ D(k) = (x * f_h)(2k).
\end{align}
The Inverse DWT~(IDWT) reconstructs the original signal by first upsampling the chosen coefficients from $A$ and $\{D_i\}$, followed by convolving them with inverse low-pass filter $g_l$ and high-pass filters $g_h$ and summing the results as below,
\begin{align}
    \hat{x}(t) = (A \uparrow 2 * g_l)(t) + (D \uparrow 2 * g_h)(t).
\end{align}
Here the upsampling operator $\uparrow 2$ inserts a zero between each pair of consecutive elements in the input signal.
DWT and IDWT enable an efficient decoupling and analysis of multi-scale components in the signals, which have been widely applied in various fields including image compression, signal denoising~\cite{wavelet1989siam}.

\subsection{Evidence of Efficiency for the Aggregator} \label{app:sum_operator}
From \tabref{tab:agg_comparison}, we can see that the sum operator in CP-MoE achieves SOTA performance over GCN and GAT aggregators in terms of Accuracy, Recall, W-F1 and C-F1. This superiority can be attributed to two factors:
(i) Learnable graph aggregators tend to excessively highlight similarities between adjacent nodes~\cite{BAGNN2022icde}. This could result in over-reliance on the link correlation induced by the often similar congestion levels within a local region in our problem. Consequently, the learnable aggregators insufficiently capture the congestion propagation patterns, which are often associated with congestion level heterophily between adjacent links. This is also corroborated by the performance differences \wrt Recall and Precision;
(ii) Under the MoE architecture, the bias in gate inputs induced by learnable aggregators compromises the stability of routing. This, in turn, adversely affects the training of multiple experts, leading to a severe decline in overall performance~\cite{robustmoecv2023iccv}.

\begin{table*}[t] \footnotesize 
\caption{The 12-step congestion prediction performance on Beijing and Shanghai datasets \wrt different spatial aggregator choices, which are used to compile short-term spatio-temporal contextual gate inputs.}
\vspace{-5pt}
\begin{tabular}{c|ccccc|ccccc}
\toprule
& \multicolumn{5}{c|}{Beijing} & \multicolumn{5}{c}{Shanghai} \\ \cline{2-11} 
\multirow{-2}{*}{Model} & Accuracy(\%) & Recall(\%) & Precision(\%) & W-F1(\%) & C-F1(\%) & Accuracy(\%) & Recall(\%) & Precision(\%) & W-F1(\%) & C-F1(\%)  \\ 
\midrule

\textbf{CP-MoE}& \textbf{0.8521} & \textbf{0.8091} & 0.7651 & \textbf{0.7590} & \textbf{0.7865} & \textbf{0.9126} & \textbf{0.6775} & 0.7288 & \textbf{0.7433} & \textbf{0.7022} \\
w/ GCN aggregator & 0.8519 & 0.7973 & 0.7723 & 0.7579 & 0.7846 & 0.9126 & 0.6651 & 0.7392 & 0.7395 & 0.7002 \\
w/ GAT aggregator & 0.8509 & 0.7859 & \textbf{0.7774} & 0.7560 & 0.7817 & 0.9121 & 0.6573 & \textbf{0.7436} & 0.7364 & 0.6978 \\

\bottomrule
\end{tabular}
\label{tab:agg_comparison}
\vspace{-3pt}
\end{table*}

\subsection{Lightweight Periodic Expert} \label{app:per_expert}
The periodic expert first concatenates the historical traffic sequences in $\mathcal{H}$ into one sequence $X^{p} \in \mathbb{R}^{T_f(N_d+N_w) \times N \times C}$ along the time dimension.
Then each slice is concatenated with its corresponding temporal embeddings to
enhance the expert's understanding of relative time positions.
Afterwards, we feed the temporally enriched historical sequence into multiple MLP layers to obtain the temporal encodings $H^{p} \in \mathbb{R}^{N \times d},$ which are further attached with learnable spatial embeddings to increase the spatial discriminability of samples. 
Finally, the combined embeddings are input to another MLP to produce the future predictions logits $\hat{P}_{per}^{t+1:t+T_f} \in \mathbb{R}^{T_f \times N \times 3}.$
In practice, the learnable spatio-temporal embeddings are shared with MAGLs as defined in \secref{subsubsec:gate_input}. 


\section{Experiment Details}
\subsection{Baseline Details} \label{app:baseline}
We detailedly introduce the compared deep learning baselines as follows.
\textbf{General STGNNs: }
(1) DCRNN~\cite{dcrnn2018iclr}: It integrates graph diffusion operation into GRU~\cite{gru2014} and make traffic prediction in an encoder-decoder manner;
(2) ASTGNN~\cite{astgnn2022tkde}: It harnesses self-attention to model the recent and periodic context and a dynamic GCN to capture spatial heterogeneity, making traffic prediction in an encoder-decoder manner;
(3) GWNet~\cite{gwnet2019ijcai}: It combines gated TCN and GCN with a learnable adaptive graph for traffic prediction;
(4) AGCRN~\cite{agcrn2020nips}: It enhances GRU with graph convolutions based on adaptive graph and proposes a node adaptive parameter learning mechanism;
(5) STID~\cite{stid2022cikm}: It leverages efficient MLP layers with learnable spatial and temporal embeddings for multivariate time series forecasting.
(6) STWave~\cite{stwave2023tkde}: It disentangles traffic data into trend and event signals, models them separately, and adaptively fuses them via attention mechanism for prediction.
(7) ST-MoE~\cite{stmoe2023cikm}: It integrates multiple STGNNs into an MoE framework. Each expert is chosen as AGCRN, which outperforms all others expert choices according to our empirical results; 
(8) STAEformer~\cite{staeformer2023cikm}: It incorporates three types of spatio-temporal adaptive embeddings to strengthen the vanilla Transformer on spatio-temporal forecasting.
\textbf{Congestion prediction methods: }
(9) DuTraffic~\cite{dutraffic2022cikm}: It is a congestion prediction model deployed at BaiduMap, where we only compared with its multi-task learning framework due to the unavailability of the visual data; 
(10) STTF~\cite{sttf2023ijcis}: It is an autoregressive spatio-temporal transformer designed for congestion prediction. 
Since baselines (9) and (10) are not open-source, we reproduced their algorithms independently. 

\subsection{Implementation Details} \label{app:imple}
The hidden dimension of \model is $D=32$. The number of MAGL layers is 2. In each MAGL layer, the number of upstream, downstream and global experts are 4, 4 and 2, respectively. Top-6 experts will be activated at a time. The weights of expert balancing losses are $\lambda_1 = 10^{-3}, \lambda_2 = 10^{-3}.$
The number of TCN layers is 2 and we extract 5-hop neighbors for spatial aggregation. The dimension of learnable embeddings is $D_l=10.$
For trend expert, we use Daubechies 1 wavelet for trend decoupling and set the head of MSA as 2.
The class distance function in ordinal regression satisfies $\phi(0,1)=1, \phi(1,2)=2.$
We divide each dataset with rate 7:1:2 along the timeline for training, validation and testing, respectively.
We use Adam Optimization with learning rate $10^{-3}$, weight decay rate $5\times 10^{-7}$, dropout rate 0.15 and early stopping for 30 epochs. 
The model is trained on a high-performance server equipped with four Intel Xeon E5-2630 V4 CPUs, 45 GB of memory, and a single Nvidia Tesla P40 GPU.



\subsection{Parameter Sensitivity Study}
We study the sensitivity of \model on five hyperparameters: 
(1) the number of upstream experts $N_{up}$ in each MAGL layer, 
(2) the number of downstream experts $N_{down}$ in each MAGL layer, 
(3) the number of activated experts $k$ for each sample in each MAGL layer,
(4) the weight of important balancing loss for MAGLs $\lambda_1$,
(5) the weight of load balancing loss for MAGLs $\lambda_2$.

\begin{figure}[ht] 
\begin{minipage}{1.0\linewidth}
\centering  
\subfigure[\small Effect of upstream expert number.]{
    \label{fig:para-up}
    \includegraphics[width=0.29\textwidth]{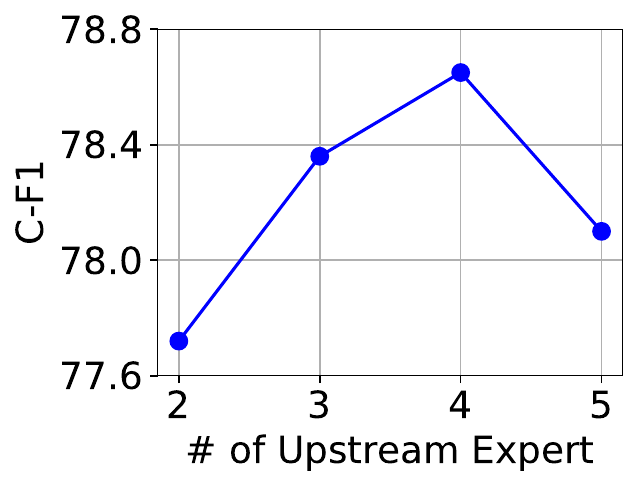}}
\quad
\subfigure[\small Effect of downstream expert number.]{
    \label{fig:para-down}
    \includegraphics[width=0.29\textwidth]{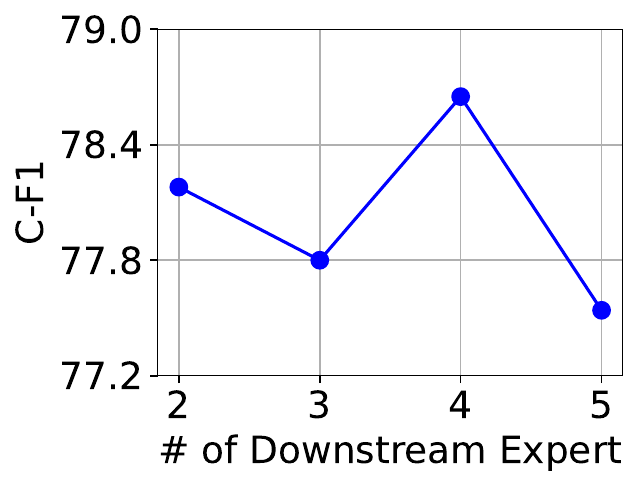}}
\quad
\subfigure[\small Effect of Top-$k$ activation.]{
    \label{fig:para-topk}
    \includegraphics[width=0.29\textwidth]{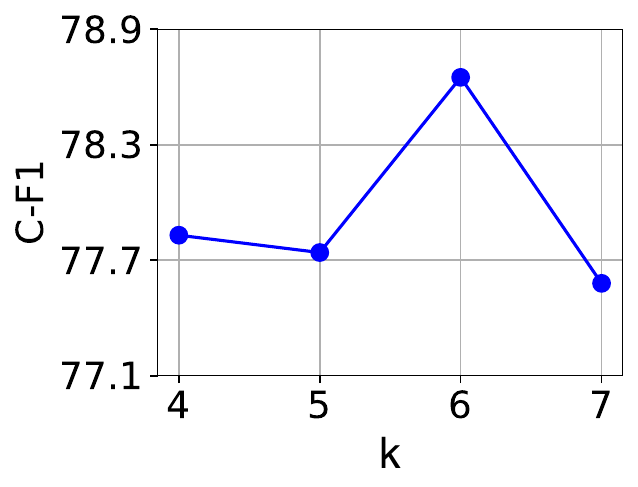}} \\

\subfigure[\small Effect of $\lambda_1$.]{
    \label{fig:para-lambda1}
    \includegraphics[width=0.29\textwidth]{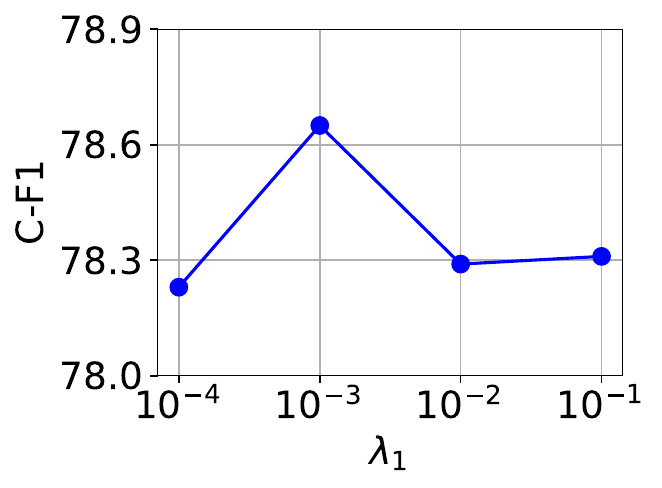}}
\subfigure[\small Effect of $\lambda_2$.]{
    \label{fig:para-lambda2}
    \includegraphics[width=0.29\textwidth]{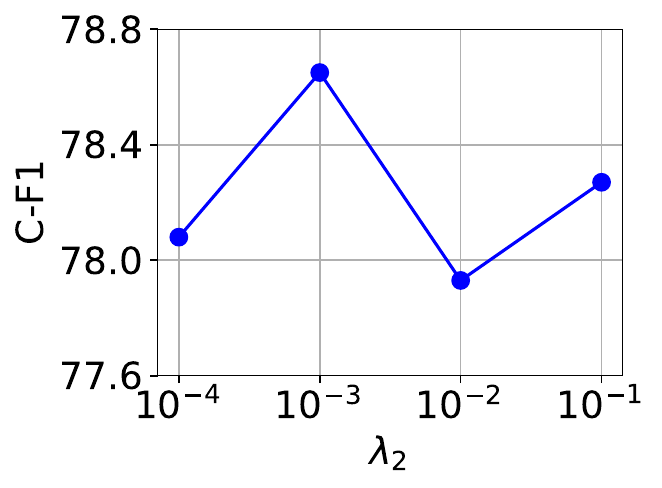}} 
    
\end{minipage}
\caption{Parameter sensitivity analysis on Beijing dataset.}
\label{fig:sens}
\end{figure}

First, we vary $N_{up}$ and $N_{down}$ from 2 to 5, respectively. As reported in \figref{fig:para-up} and \ref{fig:para-up}, the model achieves the best performance when $N_{up} = N_{down} = 4$. Intuitively, insufficient experts can limit the model's learning capability in certain scenarios, whereas too many experts can render the model hard to converge.

Second, we vary $k$ from 4 to 7 as depicted in \figref{fig:para-topk}. The model achieves the best performance when $k=6.$ Overly sparse activation may compromise prediction accuracy while overly dense activation may hinder the specialized training of experts.

Third, we vary $\lambda_1$ and $\lambda_2$ from $10^{-4}$ to $10^{-1},$ respectively. As shown in \figref{fig:para-lambda1} and \ref{fig:para-lambda2}, the best performance is achieved when $\lambda_1 = \lambda_2 = 10^{-3}.$ Intuitively, over-emphasis on expert balancing in MAGLs may interfere the specialization of different experts, whereas insufficient expert balancing can lead to suboptimal expert collaboration modes.

Overall, \model's performances \wrt C-F1 vary within an acceptable range on the Beijing dataset, demonstrating its robustness against different hyperparameters.

\eat{
\subsection{System Deployment} \label{app:deploy}
We have used Spark to build an efficient data processing pipeline, consisting of two main steps: (i) extracting link-level traffic features from ride-hailing car trajectories every minute, (ii) aggregating congestion level data for each 5-min timestep for use in CP-MoE's prediction. The deployed CP-MoE predicts future congestion on circular road links within Beijing's 5th Ring Road. These links are relevant to over 500K daily ride-hailing orders. The predictions for the upcoming hour (12 time steps) are updated every 5 minutes based on the traffic data in the previous hour.
CP-MoE and the optimal replacement time are daily updated. Once the updation is completed, the model will be pushed to online servers to offer real-time congestion prediction. For each TTE query, its congestion level features are now selected from CP-MoE's predictions. For the complete online processing pipeline of the TTE system, please refer to Appendix A.4.3 in our previous work~\cite{ieta2023kdd}.
}

\end{document}